\documentclass[10pt,twocolumn,letterpaper]{article}

\usepackage{cvpr}              
\usepackage{multicol}
\usepackage{multirow}
\usepackage{float}

\usepackage{array}
\usepackage{tabularx}
\usepackage{booktabs}
\usepackage{makecell}
\usepackage{bbding}
\usepackage{pifont}
\usepackage{wasysym}
\usepackage{amssymb}
\usepackage{soul}
\usepackage{graphicx} 
\usepackage{float} 
\usepackage{subcaption}

\usepackage{algorithm}
\usepackage{algorithmicx}
\usepackage{algpseudocode}
\usepackage{tcolorbox}
\usepackage{colortbl}
\usepackage{stfloats}
\usepackage{color,xcolor}
\definecolor{deep_red}{HTML}{C00000}
\definecolor{shallow_green}{HTML}{70AD47}
\definecolor{deep_blue}{HTML}{4472C4}

\usepackage[T1]{fontenc}
\definecolor{cvprblue}{rgb}{0.21,0.49,0.74}
\definecolor{dark_green}{rgb}{0, 0.5, 0}
\usepackage[pagebackref,breaklinks,colorlinks,allcolors=cvprblue]{hyperref}

\def\paperID{31446}
\def\confName{CVPR}
\def\confYear{2026}

\title{TIM: Temporal Decoupling with Iterative Mutual-Refinement Model for Longitudinal Radiology Report Generation}

\author{
\normalsize Yiheng Dong$^{1}$\thanks{Equal contribution.}, Yi Lin$^{2}$\footnotemark[1], Shilong Huang$^{1}$, Xiyan Yang$^{1}$, Xin Yang$^{1}$\thanks{Corresponding author.}\\
\normalsize $^{1}$ Huazhong University of Science and Technology, Wuhan, China\quad $^{2}$ HKUST, Hong Kong, China \\
{\tt\small \{yiheng, alon\_ft, xiyanyang, xinyang2014\}@hust.edu.cn}\quad {\tt\small yi.lin@connect.ust.hk}
}

\begin{document}
\maketitle
\begin{abstract}
Automatic radiology report generation (RRG) aims to translate medical images into diagnostic text, reducing radiologists' workload and standardizing clinical documentation. 
Nonetheless, existing approaches mainly focus on single-timepoint analysis and fail to capture temporal disease evolution across longitudinal examinations. 
While recent longitudinal RRG (LRRG) approaches incorporate historical data, they often combine images from different time points within a single representation space, leading to blurred semantics and inconsistent temporal reasoning.
In this work, we propose a \textbf{T}emporal Decoupling with \textbf{I}terative \textbf{M}utual-Refinement Model (TIM), a two-stage framework that explicitly decouples spatial pathology from temporal progression and iteratively refines reports through mutual feedback. Stage~I performs temporal-decoupled representation learning, separating temporal evolution patterns from disease-specific features and generating radiology reports for both prior and current studies. 
Stage~II introduces a mutual report refinement mechanism that identifies diagnostic inconsistencies within prior reports and iteratively rectifies both prior and current reports through error-sensitive feedback.
Experiments on the Longitudinal-MIMIC dataset demonstrate that TIM surpasses existing single-image and longitudinal baselines, achieving new state-of-the-art performance across both language and clinical metrics. Code is available at \url{https://github.com/yihengd/TIM}.
\end{abstract}    
\section{Introduction}
\begin{figure}[!t]
\vspace{-3mm}
\centering\includegraphics[width=\linewidth]{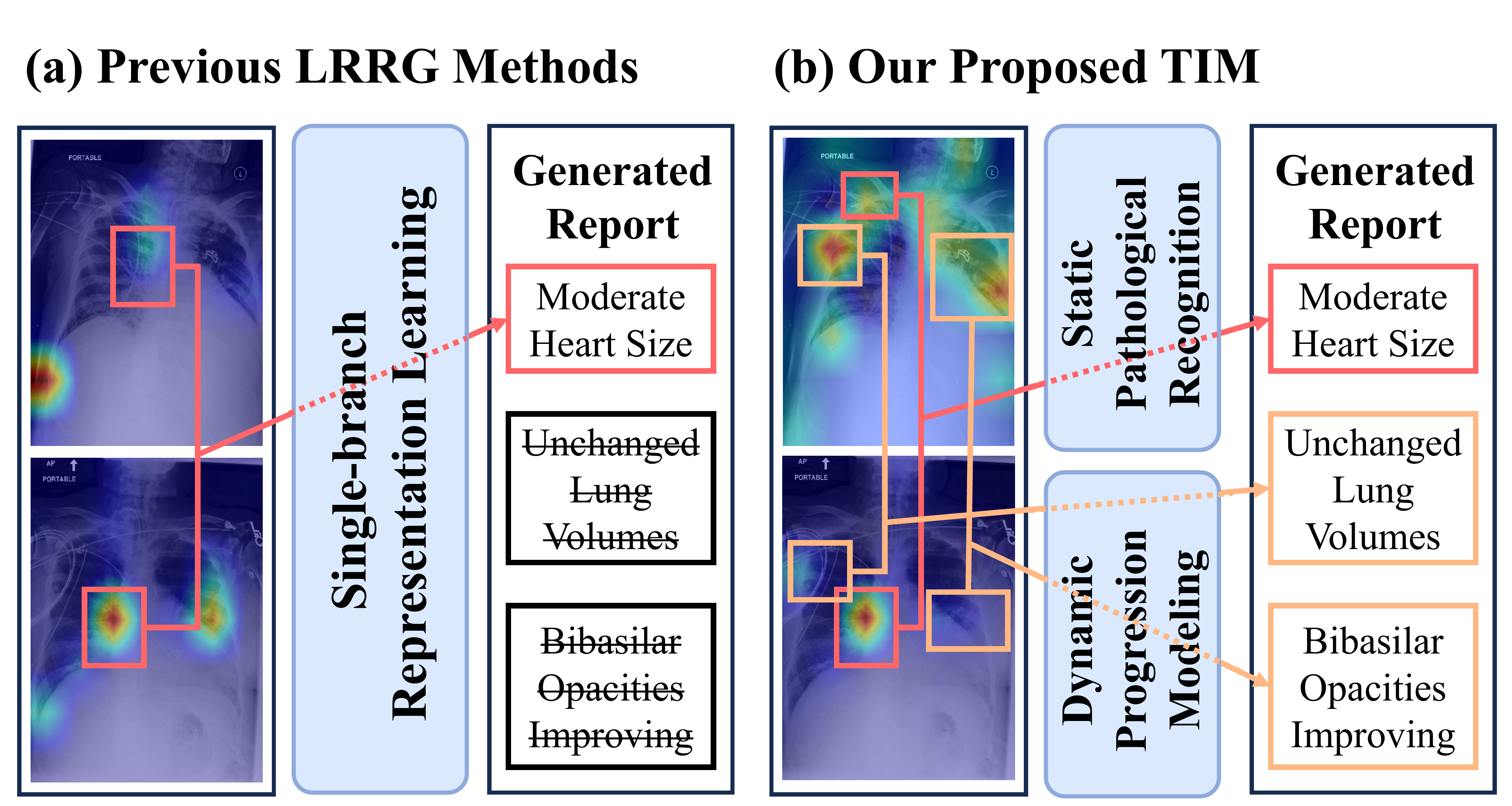}
\caption{\textbf{Comparison between TIM and previous work.} 
Previous methods typically use a single representation branch, limiting their ability to capture disease progression. In contrast, TIM decouples the longitudinal radiology report generation into \textit{static pathology recognition} and \textit{dynamic progression modeling}, allowing precise characterization of both prior and current pathological states, as well as their longitudinal evolution.}
\label{fig:figure1}
\vspace{-5mm}
\end{figure}
Radiology report generation (RRG) aims to automatically produce diagnostic descriptions from medical images, offering great potential to alleviate radiologists' workload and standardize clinical documentation. By converting images into structured diagnostic narratives, RRG systems can assist physicians in drafting reports, standardize terminology usage, and reduce human variability~\cite{johnson2019mimic, li2023dynamic}.

Recent advances in vision-language models~\cite{ji2021improving, li2023blip, liu2023llava, vinyals2015show} have enabled significant progress in RRG, achieving accurate report generation for chest X-rays~\cite{chen2020r2gen, chen2022CMN, bu2024ekagen, jin2024promptmrg, xiang2024gmod, wang2023r2gengpt, park2025dart, hou2025radar}. Despite these advances, most existing approaches focus solely on single-timepoint studies, overlooking the temporal continuity that is intrinsic to real clinical workflows~\cite{wang2024hergen}. In practice, radiologists routinely compare the current study with prior examinations to assess disease progression, treatment response, and emergence of new abnormalities~\cite{dong2025copa}. Without modeling temporal evolution, single-timepoint methods struggle to capture longitudinal progression and generate reports that are temporally inconsistent or misaligned with the patient's actual disease trajectory.

To address these limitations, several recent studies~\cite{zhu2023longitudinalmimic, wang2024hergen, nicolson2024cxrmate, liu2025mlrg, liu2025hcllm, yun2025diffrrg, yang2025stream} have explored longitudinal radiology report generation (LRRG). 
These approaches incorporate historical images and reports to produce reports encompassing both current findings and disease progression. 
While these advancements mark a significant step forward, they overlook a fundamental property of the LRRG task: the heterogeneity between pathological recognition and progression modeling. 
Pathological recognition is concerned with the spatial localization~\cite{lin2025boosting} of disease-specific patterns within a single study (e.g., pulmonary opacity), whereas progression modeling captures temporal changes across studies (e.g., interval increase or improvement). 
As shown in Fig.~\ref{fig:figure1}(a), existing methods conflate these two distinct subtasks within a single representation learning network, resulting in the blend of static disease cues and temporal evolution features. 
This entanglement obscures their unique semantics and limits the model's capacity for accurate temporal reasoning. 

In addition, existing LRRG methods treat report generation as a single-pass procedure, disregarding the mutual dependencies between prior and current reports. 
By treating each report as an independent target, these methods fail to leverage the bidirectional cues across time points. 
Consequently, they ignore two critical opportunities for enhancing diagnostic reasoning: first, identifying errors in the prior generated report can offer corrective guidance for the current report; second, refining the current report can reveal omissions or misinterpretations in the prior one. 
Without mutual refinement, such models often repeat mistakes across visits and fail to characterize disease evolution.

In this study, we propose Temporal Decoupling with Iterative Mutual-Refinement Model (TIM), a two-stage framework for longitudinal radiology report generation. 
Stage~I performs temporal-decoupled representation learning, wherein a static pathological recognition branch and a dynamic progression modeling branch independently encode disease-specific visual features and cross-visit progression representations, mitigating feature interference between spatial and temporal cues. 
Based on these representations, an LLM generates radiology reports for both prior and current examinations. 
Stage~II introduces mutual report refinement, whereby the prior and current reports serve as reciprocal references to guide the iterative refinement process. Through improved clinical accuracy and temporal coherence, this stage ultimately improves the overall quality and reliability of longitudinal radiology reporting.

Our key contributions are summarized as follows:
\begin{enumerate}[1. ]
\item We introduce TIM, a Temporal-Decoupling with Iterative Mutual-Refinement model that adopts a ``divide-and-conquer'' strategy to exploit the temporal evolution in longitudinal radiology reports.
\item We propose a temporal-decoupled representation learning strategy that separates static pathology from dynamic progression, and employs a dual-timepoint report generation mechanism to capture longitudinal disease semantics while maintaining longitudinal consistency.
\item We design a mutual report refinement module, wherein both prior and current reports are iteratively refined through mutual referencing, effectively reducing temporal error propagation.
\item Experiments on the Longitudinal-MIMIC dataset demonstrate that TIM outperforms both single-image and longitudinal baselines across natural language generation and clinical efficiency metrics.
\end{enumerate}
\section{Related Work}

\textbf{Radiology Report Generation.} RRG aims to automatically produce diagnostic reports from medical images such as chest X-rays. While this task shares similarities with image captioning~\cite{vinyals2015show, xu2015show, ji2021improving}, it is inherently more complex due to the need for precise clinical terminology and long-form structured descriptions. Early methods generally adopted an encoder-decoder framework~\cite{anderson2018bottom, cornia2020meshed, pan2020x, vinyals2015show, xu2015show}, where convolutional or transformer-based encoders extract image features and recurrent or transformer-based decoders generate textual reports. To enhance the clinical relevance of generated reports, several works~\cite{tanida2023interactive, wang2022medclip, liu2021exploring, hou2025radar, jin2024promptmrg, bu2024ekagen,fang2024aligning} incorporated domain knowledge or multimodal alignment strategies. Despite these advances, most prior works focus on single-image settings, overlooking the temporal evolution of disease progression across multiple visits.

\noindent\textbf{Longitudinal Radiology Report Generation.} 
LRRG extends traditional RRG to multi-visit scenarios, enabling the capture of disease progression and treatment response. Unlike conventional RRG models that describe only static findings, LRRG aims to model both current pathologies and longitudinal evolutions. Recent methods such as CXRMate~\cite{nicolson2024cxrmate}, Prefilling~\cite{zhu2023longitudinalmimic}, HERGen~\cite{wang2024hergen}, HC-LLM~\cite{liu2025hcllm}, and Diff-RRG~\cite{yun2025diffrrg} incorporate temporal or historical information to enhance contextual understanding. However, these approaches often mix static and temporal representations, limiting feature disentanglement. In contrast, our method employs a decoupled representation learning strategy to separate static and dynamic visual cues, enabling the model to capture both stable and evolving abnormalities for more accurate and coherent report generation.

\noindent\textbf{Self-Refinement Report Generation.}
Self-refinement has emerged as an effective strategy for improving generative models through iterative revision of outputs based on feedback~\cite{madaan2023self, welleck2022generating}, enhancing factual accuracy and linguistic coherence in natural language generation tasks~\cite{manakul2023selfcheckgpt, kumar2024training}. 
However, existing self-refinement methods for report generation~\cite{park2025dart} often fail to provide explicit and clinically meaningful feedback, resulting in limited guidance and marginal gains in clinical accuracy.
In contrast, we introduce a dual-timepoint, iterative mutual-refinement approach that goes beyond previous self-refinement by explicitly identifying errors in reports, enabling targeted correction, and substantially improving the clinical accuracy of the final output.
\section{Method}
\begin{figure*}[!t]
\centering\includegraphics[width=\textwidth]{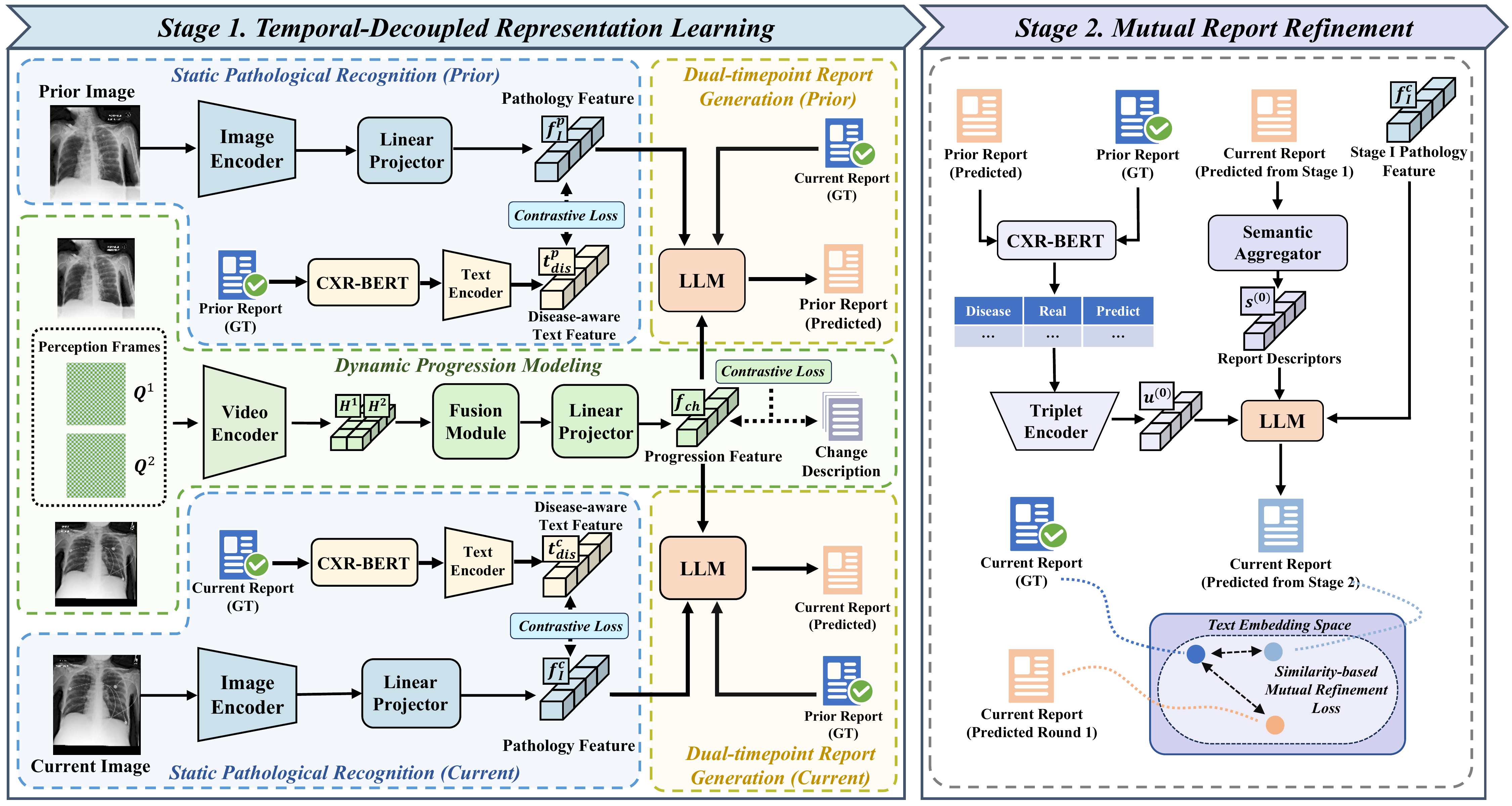}
\vspace{-3mm}
\caption{\textbf{Overview of the proposed TIM.} Our proposed model comprises two stages: (I) temporal-decoupled representation learning and (II) mutual report refinement. During Stage I, TIM generates dual-timepoint reports using static pathological recognition and dynamic progression modeling. In Stage II, TIM identifies errors in reports and iteratively performs mutual refinement, generating clinically accurate and temporally coherent reports.}
\label{fig:framework}
\vspace{-3mm}
\end{figure*}
The overall framework is shown as Fig.~\ref{fig:framework}. Our proposed TIM comprised two stages: (I) \textbf{Temporal-Decoupled Representation Learning} and (II) \textbf{Mutual Report Refinement}. In Stage~I, to mitigate feature interference and enhance the specificity for disease-aware findings and temporal evolution, TIM decouples visual representations into pathological recognition and progression modeling branches, and performs report generation for two timepoints. Stage~II further enhances clinical accuracy by introducing \textbf{Mutual Report Refinement}. Through multiple rounds of mutual refinement, TIM suppresses mistakes progressively and yields clinically faithful radiology reports.

\subsection{Problem Formulation}
Let $\mathcal{I}=\{I^p, I^c\}$ denote the pair of chest X-ray images acquired from the same patient at two different times, where ${I^p}$ and ${I^c}$ denote the prior and current images, respectively. Each image is associated with a corresponding report $R^p$ or $R^c$. In the following sections, since our framework involves report generation at two timepoints, we introduce the superscript ``$\ast$'' to uniformly denote a generic timepoint. For instance, $I^\ast$ may correspond to either the prior or current image, and the same notation applies to others.

Given a prior image–report pair $\{I^p, R^p\}$ and a current image $I^c$, our goal is to generate an accurate and detailed current report $\hat{R}^c$, which can be formally represented as:
\begin{equation}
    \mathcal{F}: (I^p, I^c, R^p) \rightarrow \hat{R}^c.
\end{equation}

\subsection{Temporal-Decoupled Representation Learning}
In Stage~I, we decouple representation learning into \textbf{Static Pathological Recognition} and \textbf{Dynamic Progression Modeling} branches, enabling the model to capture prior and current pathological states and their longitudinal evolution. Based on these representations, a \textbf{Dual-Timepoint Report Generation} mechanism is established to produce both prior and current radiology reports. 

\subsubsection{Static Pathological Recognition}
\label{sec: pathological recognition}
This branch is designed to capture the static disease-related visual representation from prior and current images independently. Given an image $I^\ast$, an image encoder $E_{img}(\cdot)$ extracts patch embeddings, which are then projected into language space through a linear projector $W_{img}$, forming a pathological feature $\mathbf{f}_i^\ast=W_{img}E_{img}(I^\ast)$. 

To couple the pathological feature with clinically faithful semantics, we align $\mathbf{f}_i^\ast$ with disease-aware textual descriptions. Specifically, each corresponding report is first annotated by CheXbert model~\cite{irvin2019chexpert} across 14 disease categories, producing four labels: \{\textit{not mentioned}, \textit{positive}, \textit{negative}, \textit{uncertain}\}. Using external medical knowledge sources (e.g., UMLS, PubMed), each mentioned disease is expanded into a concise clinical description. These descriptions and their labels are concatenated into a disease text sequence $\mathcal{T}_{dis}^\ast$ and processed by a text encoder into $\mathbf{f}_{t,dis}^\ast = E_{txt}(\mathcal{T}_{dis}^\ast)$. We then apply a contrastive learning to align image and disease-aware text features:
\begin{equation}
\mathcal{L}_{spr}^\ast=-\log\frac{\exp(sim(\mathbf{f}_i^\ast,\mathbf{f}_{t,dis}^\ast)/\tau)}{\sum_j\exp(sim(\mathbf{f}_i^\ast,\mathbf{f}_{t, dis}^{\ast,j})/\tau)},
\label{contrastive loss 1}
\end{equation}
where $sim(\cdot,\cdot)$ denotes cosine similarity function, $\tau$ is a temperature hyper-parameter, and $\mathcal{L}_{spr}^\ast$ denotes the loss of static pathological recognition branch.

\subsubsection{Dynamic Progression Modeling}
\label{sec:dpm}
This branch focuses on capturing temporal evolution between consecutive examinations by exploiting the intrinsic spatiotemporal learning capability of the video encoder. We introduce a set of learnable, randomly initialized perception frames $\{Q^1, Q^2\}$ that share the spatial resolution of chest images $\{I^p, I^c\}$ and are independently parameterized. The number of perception frames is set to 2 based on experiments in Sec.~\ref{exp:hyperparam}. These frames are inserted between the prior and current images to form a short video sequence, which is then processed by a video encoder $E_{vid}(\cdot)$:
\begin{equation}
    (\_, H^1, H^2,\_)=E_{vid}([I^p; Q^1; Q^2; I^c]),
\end{equation}
where $[\cdot;\cdot]$ denotes concatenation operation. The resulting features $\{H^1, H^2\}$ are adaptively fused by a gating module:
\begin{equation}
    H_{g}=\alpha \odot H^1 + (1-\alpha) \odot H^2 ,\quad \alpha=\sigma(W_{g}[H^1;H^2]),
\end{equation}
where $\odot$ denotes element-wise multiplication, $\sigma(\cdot)$ is the sigmoid function, and $W_g$ is a learnable weight matrix. $H_{g}$ is then projected into the language space by a linear head $W_{vid}$, resulting in a progression feature $\mathbf{f}_{pro}=W_{vid}H_{g}$.

To ensure that the learned representations reflect clinically meaningful progressions, we align $\mathbf{f}_{pro}$ with text embeddings $\mathbf{f}_{t,pro}=E_{txt}(\mathcal{T}_{pro})$ derived from progression-related report sentences $\mathcal{T}_{pro}$:
\begin{equation}
\mathcal{L}_{dpm}=-\log\frac{\exp(sim(\mathbf{f}_{pro},\mathbf{f}_{t,pro})/\tau)}{\sum_j\exp(sim(\mathbf{f}_{pro},\mathbf{f}_{t,pro}^j)/\tau)},
\end{equation}
where $\mathcal{L}_{dpm}$ denotes the loss of dynamic progression modeling branch. This alignment encourages the model to focus on true pathological evolution rather than irrelevant imaging variations such as positioning changes.

\subsubsection{Dual-Timepoint Report Generation}
To better capture temporal evolution, Stage~I generates both prior and current reports rather than predicting only the current one. This dual-timepoint formulation enables the model to infer the current report based on detected progression, while simultaneously using progression to reconstruct the prior report. Such a bidirectional reasoning process encourages the progression representation to encode disease evolution more accurately and consistently across time. We utilize a frozen large language model $LLM$ as the report generator, taking embeddings and a textual prompt as input:
\begin{equation}
    \hat{R}^\ast = LLM(\mathbf{f}_i^\ast, \mathbf{f}_{pro}, R^{ref}, \mathcal{P}^\ast)
    \label{stage 1 report generation},
\end{equation}
where $\hat{R}^\ast$ denotes the generated report at time $\ast$ ($p$ or $c$), $\mathbf{f}_i^\ast$ and $\mathbf{f}_{pro}$ are the pathology and progression embedding, $R^{ref}$ is the reference report from the opposite time ($R^p$ for current generation and $R^c$ for the prior one) and $\mathcal{P}^\ast$ is a textual prompt indicating the timepoint of the generated report and assisting LLM to understand the task objective. 

During training, both timepoint generations are optimized with the same autoregressive objective:
\begin{equation}
    \mathcal{L}_{gen}^\ast=-\sum_{m=1}^M\log P_{\theta}\left(r_{m}^{*} \mid r_{<m}^{*}, \mathbf{f}_{i}^{*}, \mathbf{f}_{pro}, R^{ref}, \mathcal{P}^\ast\right),
    \label{autoregression loss 1}
\end{equation}
where $M$ denotes the total number of generated tokens and $\hat{r}_m^\ast$ denotes the $m^{th}$ text token.

The overall training objective of Stage~I is:
\begin{equation}
    \mathcal{L}_{stage1}=\mathcal{L}_{gen}^c+\mathcal{L}_{gen}^p+\lambda_1(\mathcal{L}_{spr}^c+\mathcal{L}_{spr}^p)+\lambda_2\mathcal{L}_{dpm},
    \label{stage1 loss}
\end{equation}
where $\lambda_1$ and $\lambda_2$ denotes the balancing coefficient.

\subsection{Mutual Report Refinement}
Although Stage~I carries out a detailed feature modeling process, the generated reports may still contain inaccurate or inconsistent disease descriptions, particularly for subtle abnormalities. We further observe that about 70\% of such errors in prior reports reappear in the current one owing to the temporal consistency across visits. Stage~II introduces a mutual refinement mechanism to reduce this issue. It identifies these errors by comparing the generated prior report with its ground truth, and uses them as refinement guidance for a second-round report generation. To clarify notation, throughout this stage, we use the superscript $(j)$ to denote the variable obtained after the $j^{th}$ refinement iteration.

Following the Stage~I paradigm (Eq.~(\ref{stage 1 report generation})), the model first produces initial current and prior reports $\hat{R}^{c,(0)}$ and $\hat{R}^{p,(0)}$. The generated prior report is compared with its ground truth using CheXbert~\cite{irvin2019chexpert}, yielding a set of error triplets $\mathcal{D}^{(0)}=\{(d_k,z_k^p,\hat{z}_k^{p,(0)})\}_{k=1}^K$, where $d_k$ denotes the $k^{th}$ disease, $z_k^p/\hat{z}_k^{p,(0)}$ respectively denote the ground-truth and predicted labels, and $K$ denotes the number of mistakenly predicted diseases. These triplets are then converted into high-dimensional vectors $\mathbf{u}^{(0)}$ using a triplet encoder $E_{tri}(\cdot)$ composed of concatenated one-hot vectors followed by an MLP. In parallel, to capture the core diagnostic information from the initially generated current report $\hat{R}^{c,(0)}$, we introduce a semantic aggregator $\Gamma(\cdot)$. This module applies attention mechanism over the embedding of $\hat{R}^{c,(0)}$ and produces compact report discriptors $\mathbf{s}^{(0)}=\Gamma(\hat{R}^{c,(0)})$. 

These components, along with the current pathology feature $\mathbf{f}_i^c$ from Stage~I and a textual prompt $\mathcal{P}^{rf}$, constitute the input of the second-round generation to re-evaluate the misclassified diseases and refine the current report:
\begin{small}
\begin{equation}
    \hat{R}^{c,(1)}=LLM(\mathbf{f}_i^c,\mathbf{u}^{(0)},\mathbf{s}^{(0)},\mathcal{P}^{rf}),
    \label{stage 2 report generation}
\end{equation}
\end{small}
where prompt $\mathcal{P}^{rf}$ plays the role of instructing LLM to attend to the errors and carry out refinement.

\noindent\textbf{Training.}
Training proceeds under two objectives. The first is the autoregressive generation loss, similar to Eq.~(\ref{autoregression loss 1}):
\begin{equation}
    \mathcal{L}_{gen}^{c,(1)}=-\sum_{m=1}^M\log P_{\theta}\left(r_{m}^{c,(1)} \mid r_{<m}^{c,(1)}, \mathbf{f}_i^c,\mathbf{u}^{(0)},\mathbf{s}^{(0)},\mathcal{P}^{rf}\right).
\end{equation}
To ensure the refinement leads to effective improvement, we compare the similarity of both first- and second-round reports to the ground truth in a text embedding space using cosine similarity $s(\cdot, \cdot)=sim(E_{txt}(\cdot), E_{txt}(\cdot))$:
\begin{equation}
    \mathcal{L}_{\Delta}=-\log \sigma\left(\beta\left[s(\hat{R}^{c,(1)}, R^{c})-s(\hat{R}^{c,(0)}, R^{c})\right]\right),
    \label{refinement loss}
\end{equation}
where $\sigma(\cdot)$ represents the sigmoid function and $\beta$ is a scaling hyper-parameter. This term encourages $s(\hat{R}^{c,(1)}, R^{c}) > s(\hat{R}^{c,(0)}, R^{c})$, driving the refined report to get closer to the ground truth. The full objective is:
\begin{equation}
    \mathcal{L}_{stage2}=\mathcal{L}_{gen}^{c,(1)}+\lambda_{\Delta}\mathcal{L}_{\Delta},
    \label{stage2 loss}
\end{equation}
with $\lambda_{\Delta}$ balances the two terms. Only $\Gamma(\cdot)$ and $E_{tri}(\cdot)$ are optimized while all Stage~I modules are frozen.

\begin{algorithm}[t]
\caption{Test-time Iterative Mutual-Refinement}
\label{alg:self_refine}
\textbf{Modules: }

\quad$\text{CheXbert}(\cdot)$: disease label extractor, 

\quad$LLM(\cdot)$: large language model,

\quad$\Gamma(\cdot)$: semantic aggregator,

\quad$E_{tri}(\cdot)$: triplet encoder

\textbf{Inputs: }

\quad current pathology feature $\mathbf{f}_i^c$, 

\quad prior pathology feature $\mathbf{f}_i^p$, 

\quad progression feature $\mathbf{f}_{pro}$, 

\quad prior ground truth report $R^{p}$, 

\quad textual prompts \{$\mathcal{P}^{c}$, $\mathcal{P}^{p}$, $\mathcal{P}^{rf}$\}

\textbf{Output: }refined report $\hat{R}^{c,(N)}$
\begin{algorithmic}[1]
\Statex \textit{\textcolor{gray}{//~Initialize the current report}}
\State $\hat{R}^{c,(0)} = LLM(\mathbf{f}_i^c, \mathbf{f}_{pro}, R^{p}, \mathcal{P}^{c})$;
\For{$j = 0$ to $N-1$} 
\Statex \hspace{1.2em} \textit{\textcolor{gray}{// Generate the prior report using current report}}
\State $\hat{R}^{p,(j)} = LLM\big(\mathbf{f}_i^p, \mathbf{f}_{pro}, \hat{R}^{c,(j)}, \mathcal{P}^{p}\big)$;

\State $\mathcal{D}^{(j)} = \text{CheXbert }\!\big(\hat{R}^{p,(j)}\big) \text{ vs. } \text{CheXbert }\!\big(R^{p}\big)$

\State $\mathbf{u}^{(j)} = E_{tri}\big(\mathcal{D}^{(j)}\big)$

\State $\mathbf{s}^{(j)} = \Gamma\!\big(\hat{R}^{c,(j)}\big)$
\Statex \hspace{1.2em} \textit{\textcolor{gray}{// Iteratively update the current report}}
\State $\hat{R}^{c,(j+1)} = LLM\big(\mathbf{f}_i^c, \mathbf{u}^{(j)}, \mathbf{s}^{(j)}, \mathcal{P}^{rf}\big)$ 

\EndFor
\State \Return final refined report $\hat{R}^{c,(N)}$
\end{algorithmic}
\end{algorithm}
\noindent\textbf{Inference.} 
At test time, TIM performs a closed-loop mutual refinement by alternating between Stage~I generation and Stage~II correction, as shown in Algorithm~\ref{alg:self_refine}. We begin with Stage~I to obtain $\hat{R}^{c,(0)}$, and using it as the reference report to generate $\hat{R}^{p,(0)}$ (cf. Eq.~(\ref{stage 1 report generation})). Next, we invoke Stage~II to correct and produce a new current report $\hat{R}^{c,(1)}$ (cf. Eq.~(\ref{stage 2 report generation})). The refined current report $\hat{R}^{c,(1)}$ is then fed back to Stage~I as the new reference for regenerating a new prior report $\hat{R}^{p,(1)}$. With $\{\hat{R}^{c,(1)}, \hat{R}^{p,(1)}\}$ in hand, we re-enter Stage~II: compute error triplets $\mathcal{D}^{(1)}$ with $(\hat{R}^{p,(1)}~vs.~R^p)$, update $\mathbf{s}^{(1)}=\Gamma(\hat{R}^{c,(1)})$, $\mathbf{u}^{(1)}=E_{tri}(\mathcal{D}^{(1)})$, and generate a further-refined current report $\hat{R}^{c,(2)}$. This Stage~I $\leftrightarrow$ Stage~II alternation proceeds for a fixed number of iterations $N$ (set to 3 based on Sec.~\ref{exp:hyperparam}). Ultimately, the mutual iterative refinement between the two stages yields progressively improved and clinically faithful current reports.
\section{Experiment}
\subsection{Dataset and pre-processing}
\textbf{Longitudinal MIMIC-CXR.} 
\label{dataset}
Experiments are conducted on the Longitudinal-MIMIC dataset~\cite{zhu2023longitudinalmimic} derived from the MIMIC-CXR~\cite{johnson2019mimic}, a large-scale dataset of de-identified chest radiographs paired with corresponding reports. 
Longitudinal-MIMIC organizes multiple visits per patient, with each sample comprising a tuple of \{previous image, previous report, current image, current report\}. 
Following the official split of MIMIC-CXR, this dataset includes 26,156 patients and 92,374 samples for training, 203 patients and 737 samples for validation, and 266 patients and 2,058 samples for testing.
We used Qwen3-plus~\cite{Yang2025Qwen3} to extract progression descriptions from reports, which are employed in dynamic progression modeling in Sec~\ref{sec:dpm}. 

\begin{table*}[]
\centering
\caption{\textbf{Experimental results of our model and previous methods on Longitudinal-MIMIC.} We report the natural language generation (NLG) metrics, including BLEU-n~(B-n), METEOR~(MTR), and ROUGE-L~(R-L), and clinical efficiency~(CE) metrics, including precision~(P), recall~(R), and F1-score. The best results are bolded, and the second-best results are underlined.}
\renewcommand{\arraystretch}{1.1}
\resizebox{\linewidth}{!}{%
\begin{tabular}{cccccccccccc}
\toprule[1pt] 
\multirow{2}{*}{\textbf{Input}} &
  \multirow{2}{*}{\textbf{Method}} &
  \multirow{2}{*}{\textbf{Venue}} &
  \multicolumn{6}{c}{\textbf{NLG Metrics}} &
  \multicolumn{3}{c}{\textbf{CE Metrics}} \\ 
  \cmidrule(r){4-9} \cmidrule(r){10-12} 
                        &           &               & B-1 & B-2 & B-3 & B-4      & MTR & R-L     & P & R & F1    \\ \hline \hline
\multirow{7}{*}{\begin{tabular}[c]{@{}c@{}}Single\\ Image\end{tabular}} & R2Gen~\cite{chen2020r2gen}     & EMNLP 2020    & 0.308  & 0.190  & 0.126  & 0.089       & 0.124  & 0.266       & 0.457     & 0.290  & 0.355 \\
                        & R2GenCMN~\cite{chen2022CMN}  & ACL 2021 & 0.328  & 0.203  & 0.135  & 0.096       & 0.133  & 0.273       & 0.512     & 0.359  & 0.422 \\
                        & R2GenGPT~\cite{wang2023r2gengpt}  & Meta-Rad 2023 & 0.396  & 0.243  & 0.161  & 0.108       & 0.155  & 0.260       & 0.495     & 0.385  & 0.404 \\
                        & PromptMRG~\cite{jin2024promptmrg} & AAAI 2024     & 0.384  & 0.229  & 0.149  & 0.104       & 0.146  & 0.261       & 0.517     & 0.453  & 0.439 \\
                        & EKAGen~\cite{bu2024ekagen}    & CVPR 2024     & 0.405  & 0.245  & 0.156  & 0.105       & 0.154  & 0.264      & 0.472      & 0.404  & 0.411 \\
                        & GMoD~\cite{xiang2024gmod}      & MICCAI 2024   & 0.378  & 0.234  & 0.155  & 0.107       & 0.162  &  0.276     & 0.496      & 0.429  & 0.460 \\
                        & RADAR~\cite{hou2025radar}     & ACL 2025      & 0.412  & 0.242  & 0.162  & 0.114       & 0.155  & 0.257      & 0.448      & 0.436  & 0.417  \\ \hline
\multirow{9}{*}{\begin{tabular}[c]{@{}c@{}}Longitudinal\\ Images\end{tabular}} 
                        & Prefilling~\cite{zhu2023longitudinalmimic}    & MICCAI 2023    & 0.343    & 0.210    & 0.141       & 0.100    & 0.137    & 0.274    & 0.506    & 0.364    & 0.423 \\
                        & HERGen~\cite{wang2024hergen}    & ECCV 2024     & 0.389  & 0.242  & 0.163  & 0.117       & 0.155  & \underline{0.282}       & 0.421     & 0.289  & 0.295 \\
                        & STREAM~\cite{yang2025stream}    & TMI 2025    & 0.394    & 0.237    & 0.144       & 0.104    & 0.143    & 0.261    & 0.472    & 0.428    & 0.411 \\
                        & MLRG~\cite{liu2025mlrg}    & CVPR 2025    & 0.416    & 0.252    & 0.157       & 0.114 & 0.158    & 0.264    & 0.507    & 0.425    & 0.418 \\
                        & LLM-RG4~\cite{wang2025llmrg4}      & AAAI 2025     & 0.417  & 0.240  & 0.155  & 0.115       & 0.147  & 0.257       & 0.498     & 0.441  & 0.436 \\
                        & HC-LLM~\cite{liu2025hcllm}    & AAAI 2025     & 0.404  & 0.247  & 0.164  & 0.116       & 0.163  & 0.271       & 0.488     & 0.415  & 0.448 \\
                        & Diff-RRG~\cite{yun2025diffrrg}  & MICCAI 2025   & 0.405  & 0.251  & 0.169  & \underline{0.120} & 0.164  & 0.276 & 0.528     & 0.430  & 0.474 \\ \cline{2-12} 
 &
  TIM-stage I &
  - &
  \underline{0.421} &
  \underline{0.259} &
  \underline{0.171} &
  0.118 &
  \underline{0.172} &
  0.275 &
  \underline{0.537} &
  \underline{0.466} &
  \underline{0.483} \\
 &
  TIM-stage II &
  - &
  \textbf{0.430} &
  \textbf{0.265} &
  \textbf{0.179} &
  \textbf{0.124} &
  \textbf{0.185} &
  \textbf{0.287} &
  \textbf{0.563} &
  \textbf{0.505} &
  \textbf{0.511} \\ 
  \bottomrule[1pt] 
\end{tabular}%
}
\label{table: compare with sota}
\vspace{-2mm}
\end{table*}
\subsection{Experiment Settings}
\label{experiment settings}
\textbf{Evaluation Metrics.} We evaluated the performances of our model using natural language generation (NLG) metrics and clinical efficacy (CE) metrics. For NLG evaluation, we employed BLEU-n~\cite{papineni2002bleu}, ROUGE-L~\cite{lin2004rouge}, and METEOR~\cite{banerjee2005meteor}. For CE metrics, following \cite{chen2020r2gen, chen2022CMN, jin2024promptmrg}, precision, recall, and F1-score are included, which are computed by mapping reports to 14 disease labels using CheXbert~\cite{irvin2019chexpert}.

\noindent\textbf{Implementation Details.}
In this study, all experiments are conducted on 2 NVIDIA RTX 5880 GPUs (48 GB each). We use Swin Transformer~\cite{liu2021swin} as the image encoder, visual branch of X-CLIP~\cite{Ni2022XCLIP} as the video encoder, text branch of CLIP~\cite{radford2021clip} as the text encoder, and LLaMA2-7B~\cite{Touvron2023Llama} as the large language model. 
For Stage~\uppercase\expandafter{\romannumeral1}, the model is trained for three epochs with a batch size of 4, using the Adam optimizer and a learning rate of $1\times10^{-4}$. For Stage~\uppercase\expandafter{\romannumeral2}, all Stage~\uppercase\expandafter{\romannumeral1} modules are frozen and only the semantic aggregator and the triplet encoder are optimized for one epoch with the same optimizer and batch size, and a learning rate of $5\times10^{-5}$. The balancing coefficients are set to $\lambda_1=0.5$, $\lambda_2=0.1$ in Eq.~(\ref{stage1 loss}) and $\lambda_\Delta=0.1$ in Eq.~(\ref{stage2 loss}), and the scaling parameter is set to $\beta=5$ in Eq.~(\ref{refinement loss}). During Stage~\uppercase\expandafter{\romannumeral2} inference, the number of refinement iterations is set to 3.

\subsection{Comparison with State-of-the-art Methods}
As shown in Table~\ref{table: compare with sota}, we compare our model with several recent approaches. Our framework achieves new state-of-the-art performance across all evaluation metrics, outperforming both single-input and longitudinal-input methods. Specifically, compared with single-image models, our method improves by 0.018 on BLEU-4, and 0.030 on ROUGE-L over RADAR~\cite{hou2025radar}, as well as 0.051 on F1-score over GMoD~\cite{xiang2024gmod}. Furthermore, relative to longitudinal-input models, our approach further achieves gains of 0.025 on BLEU-1, 0.011 on ROUGE-L, and 0.037 on F1-score over Diff-RRG~\cite{yun2025diffrrg}, demonstrating that effectively leveraging temporal image sequences together with report information substantially enhances both the linguistic quality and clinical accuracy of generated radiology reports.

\subsection{Ablation Study}
\begin{table*}[]
\centering
\caption{\textbf{Ablation study of Stage~\uppercase\expandafter{\romannumeral1}}. DPM: Decoupled Bilateral Representation Modeling. DRG: Dual-timepoint Report Generation mechanism. $\mathcal{L}_{spr}$: Static Pathology Recognition loss, $\mathcal{L}_{dpm}$: Dynamic Progression Modeling loss. The best results are in bold.}
\renewcommand{\arraystretch}{1}
\resizebox{\linewidth}{!}{
\begin{tabular}{cccccccccccccc}
\toprule[1pt] 
\multirow{2}{*}{\textbf{Model}} &
  \multirow{2}{*}{\textbf{DPM}} &
  \multirow{2}{*}{\textbf{DRG}} &
  \multirow{2}{*}{$\mathcal{L}_{spr}$} &
  \multirow{2}{*}{$\mathcal{L}_{dpm}$} &
  \multicolumn{6}{c}{\textbf{NLG Metrics}} &
  \multicolumn{3}{c}{\textbf{CE Metrics}} \\ 
  \cmidrule(r){6-11} \cmidrule(r){12-14}  
         &       &       &       &       & B-1 & B-2 & B-3 & B-4 & MTR & R-L & P & R & F1    \\ \hline \hline
Base & \ding{55} & \ding{55} & \ding{55} & \ding{55} & 0.396  & 0.233  & 0.151  & 0.108  & 0.155  & 0.260   & 0.495     & 0.385  & 0.404 \\
(a)      & \ding{51} & \ding{55} & \ding{55} & \ding{55} & 0.414  & 0.241  & 0.158  & 0.110  & 0.156  & 0.266   & 0.508     & 0.422  & 0.427 \\
(b)      & \ding{51} & \ding{51} & \ding{55} & \ding{55} & 0.416  & 0.245  & 0.164  & 0.114  & 0.161  & 0.271   & 0.509     & 0.452  & 0.446 \\ 
(c)      & \ding{51} & \ding{51} & \ding{51} & \ding{55} & 0.419  & 0.250  & 0.167  & 0.117  & 0.165  & 0.270   & 0.521     & \textbf{0.468}  & 0.460 \\
(d)      & \ding{51} & \ding{51} & \ding{55} & \ding{51} & 0.418  & 0.251  & 0.169  & 0.117  & 0.170  & 0.271   & 0.517     & 0.451  & 0.451 \\
(e) &
  \ding{51} &
  \ding{51} &
  \ding{51} &
  \ding{51} &
  \textbf{0.421} &
  \textbf{0.259} &
  \textbf{0.171} &
  \textbf{0.118} &
  \textbf{0.172} &
  \textbf{0.275} &
  \textbf{0.537} &
  0.466 &
  \textbf{0.483} \\ 
  \toprule[1pt] 
\end{tabular}
}
\label{table: ablation1}
\vspace{-1mm}
\end{table*}
\textbf{Stage~\uppercase\expandafter{\romannumeral1}.} Table~\ref{table: ablation1} presents an ablation study result assessing the contributions of each key component in Stage~\uppercase\expandafter{\romannumeral1}, including the dynamic progression modeling (DPM) branch, dual-timepoint report generation (DRG) mechanism, $\mathcal{L}_{spr}$ loss of static pathological recognition branch for disease semantic alignment, and $\mathcal{L}_{dpm}$ loss of dynamic progression modeling for progression representation alignment. 

Starting from the baseline model, which uses only the current images for visual feature extraction and only produces current reports, it shows limited ability to capture longitudinal information and clinical semantics. Adding the DPM branch (model (a)) improves all metrics (e.g., BLEU-1 from 0.396 to 0.414, F1-score from 0.404 to 0.422), demonstrating the importance of temporal feature modeling for understanding disease progression. Incorporating the DRG mechanism (model (b)) boosts BLEU-4 and F1 to 0.114 and 0.446, as dual-time generation encourages temporal coherence. Introducing the disease semantic alignment loss $\mathcal{L}_{spr}$ (model (c)) enforces consistency between static visual features and disease-aware text, raising F1-score to 0.460. Similarly, the progression alignment loss $\mathcal{L}_{dpm}$ (model (d)) aligns temporal features with progression descriptions, increasing BLEU-4 to 0.117. Finally, combining all modules (model (e)) achieves the highest overall performance across all metrics. These results confirm that each component contributes synergistically, improving both language quality and clinical accuracy.

\noindent \textbf{Stage~\uppercase\expandafter{\romannumeral2}.} 
\begin{table}[]
\centering
\caption{\textbf{Ablation study of Stage~\uppercase\expandafter{\romannumeral2}}. 
PDI: Prior Difference Information, $\mathcal{L}_\Delta$: Mutual-Refinement loss. The best results are in bold.}
\large
\renewcommand{\arraystretch}{1.1} 
\resizebox{\columnwidth}{!}{  
\begin{tabular}{ccccccccc}
\toprule[1.2pt]
\multirow{2}{*}{\textbf{PDI}} & \multirow{2}{*}{\textbf{$\mathcal{L}_\Delta$}} & \multicolumn{3}{c}{\textbf{NLG Metrics}} & \multicolumn{3}{c}{\textbf{CE Metrics}} \\ \cmidrule(r){3-5} \cmidrule(r){6-8} 
      &       & B-1   & B-4   & R-L   & P     & R     & F1    \\ 
      \hline \hline
\ding{55} & \ding{55} & 0.419 & 0.118 & 0.269 & 0.522 & 0.452 & 0.453 \\
\ding{51} & \ding{55} & 0.424 & 0.121 & 0.274 & 0.551 & 0.495 & 0.487 \\
\ding{51} & \ding{51} & \textbf{0.430} & \textbf{0.124} & \textbf{0.287} & \textbf{0.563} & \textbf{0.505} & \textbf{0.511} \\ 
\bottomrule[1.2pt] 
\end{tabular}%
}
\label{table: ablation2}
\vspace{-2mm}
\end{table}
To evaluate the proposed components in Stage~\uppercase\expandafter{\romannumeral2}, we conduct an ablation study on two key factors, including the prior difference information (PDI) as input, and mutual-refinement loss $\mathcal{L}_\Delta$. The results are summarized in Table~\ref{table: ablation2}.

Starting from the base model, where LLM receives only the image feature and the generated report from Stage~\uppercase\expandafter{\romannumeral1}, the performance remains limited with BLEU-1 of 0.419 and F1-score of 0.483. This indicates that without additional feedback or refinement objectives, the model struggles to correct its own diagnostic biases. When the prior difference input is introduced, the model gains refinement guidance from discrepancies between the generated and the GT prior reports. This addition yields notable improvements, particularly in CE metrics (e.g., precision of 0.551 and F1-score of 0.487), suggesting that feedback from errors can guide the model to recognize and correct misclassified diseases better. Further introducing the $\mathcal{L}_\Delta$ leads to the best overall performance, including BLEU-4 of 0.124 and F1-score of 0.511, confirming its role of ensuring refinement towards semantically meaningful corrections rather than arbitrary rewriting. 

\begin{figure*}[!t]
\centering\includegraphics[width=\linewidth]{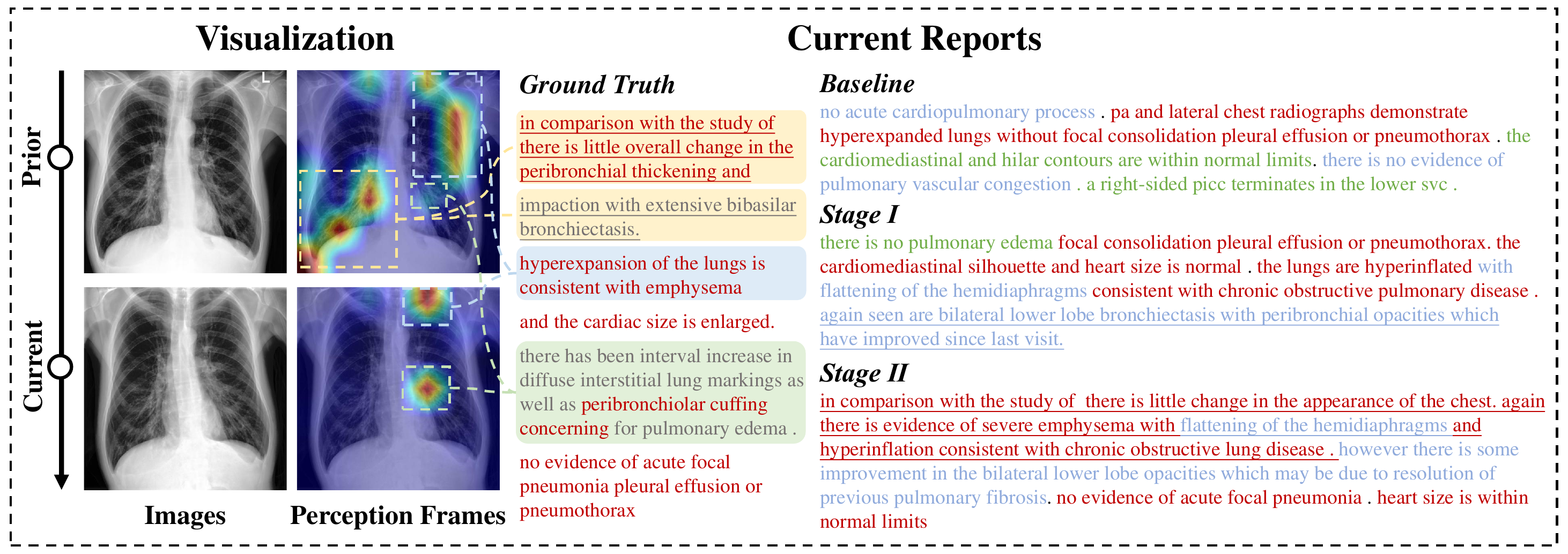}
\caption{\textbf{Case study.} On the left part, we connect the attention areas of the perception frames to the corresponding sentences in the GT report. On the right part, the red, blue, and green fonts respectively indicate statements that are consistent with, absent from, and conflicting with GT. Besides, the underlined sentences describe progression and change.}
\label{fig:case_study}
\vspace{-3mm}
\end{figure*}
\subsection{Qualitative Results}
\textbf{Case Study. }To assess the contribution of the perception frames in Stage~\uppercase\expandafter{\romannumeral1}, we visualize the captured attention regions when processing consecutive images. As shown in the left part of Fig.~\ref{fig:case_study}, the first perception frame exhibits broader attention over the image, covering extensive regions that may develop abnormalities, whereas the second one focuses on fewer but more precise regions that directly correspond to the actual lesion areas in the current image. 

We also present the generated reports from the baseline model and from TIM's Stage~\uppercase\expandafter{\romannumeral1} and \uppercase\expandafter{\romannumeral2} in Fig.~\ref{fig:case_study}. The red, blue, and green fonts respectively indicate statements that are consistent with, absent from, and conflicting with GT, and the underlined sentences describe progressions. As shown in the figure, the baseline model fails to capture any mention of evolution and produces several incorrect diagnoses. For example, it incorrectly describes the hilar contours as within normal limits, whereas the GT report indicates enlargement. On the contrary, our Stage~\uppercase\expandafter{\romannumeral1} includes descriptions of progression but still contains minor inaccuracies. These errors are effectively corrected in Stage~\uppercase\expandafter{\romannumeral2} report, which covers the most comprehensive description of the disease and its progression.

\begin{figure}[!t]
\centering

\subfloat[]{
\begin{minipage}[t]{0.48\linewidth}
\centering
\centerline{\includegraphics[width=\textwidth]{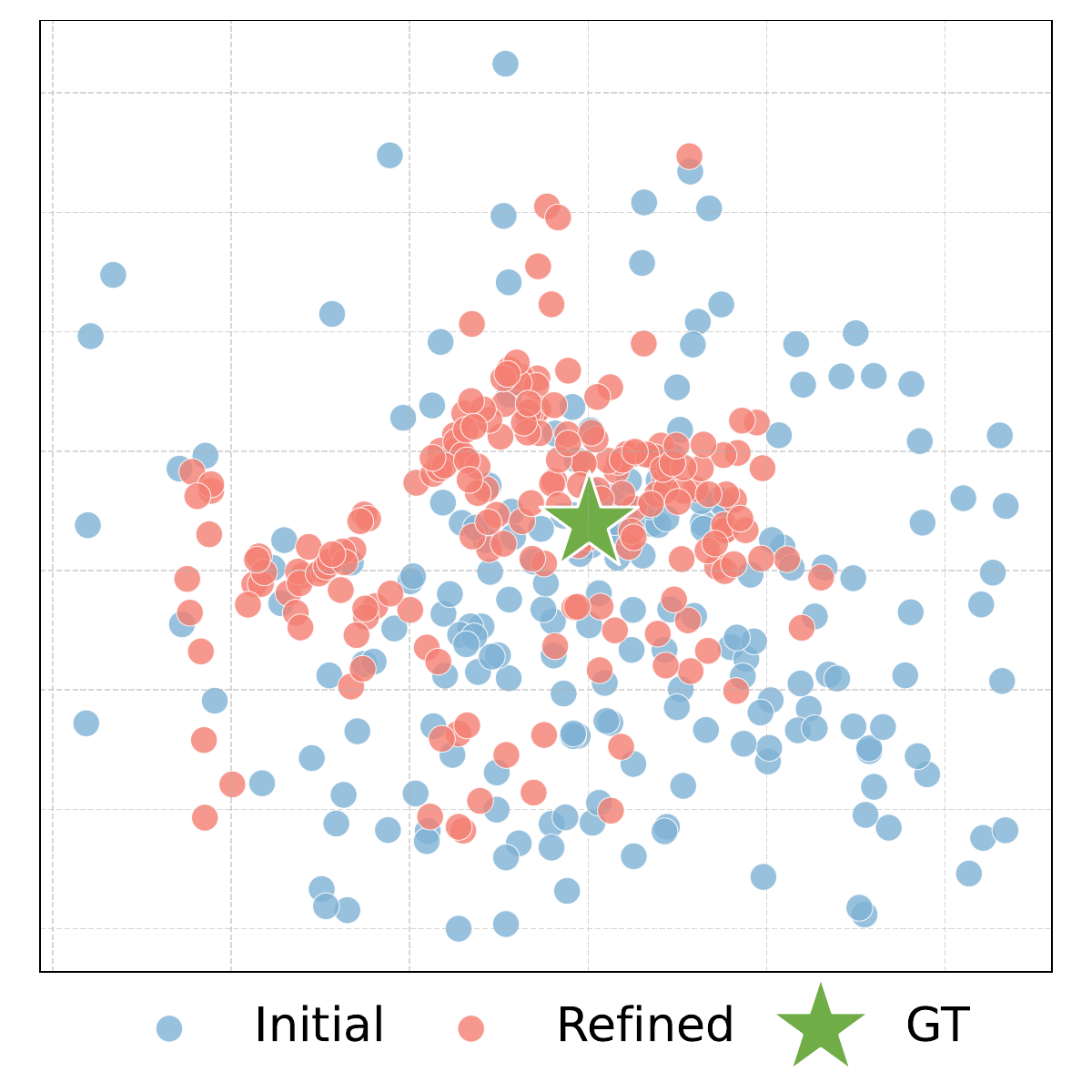}}
\end{minipage}
}
\subfloat[]{
\begin{minipage}[t]{0.48\linewidth}
\centering
\centerline{\includegraphics[width=\textwidth]{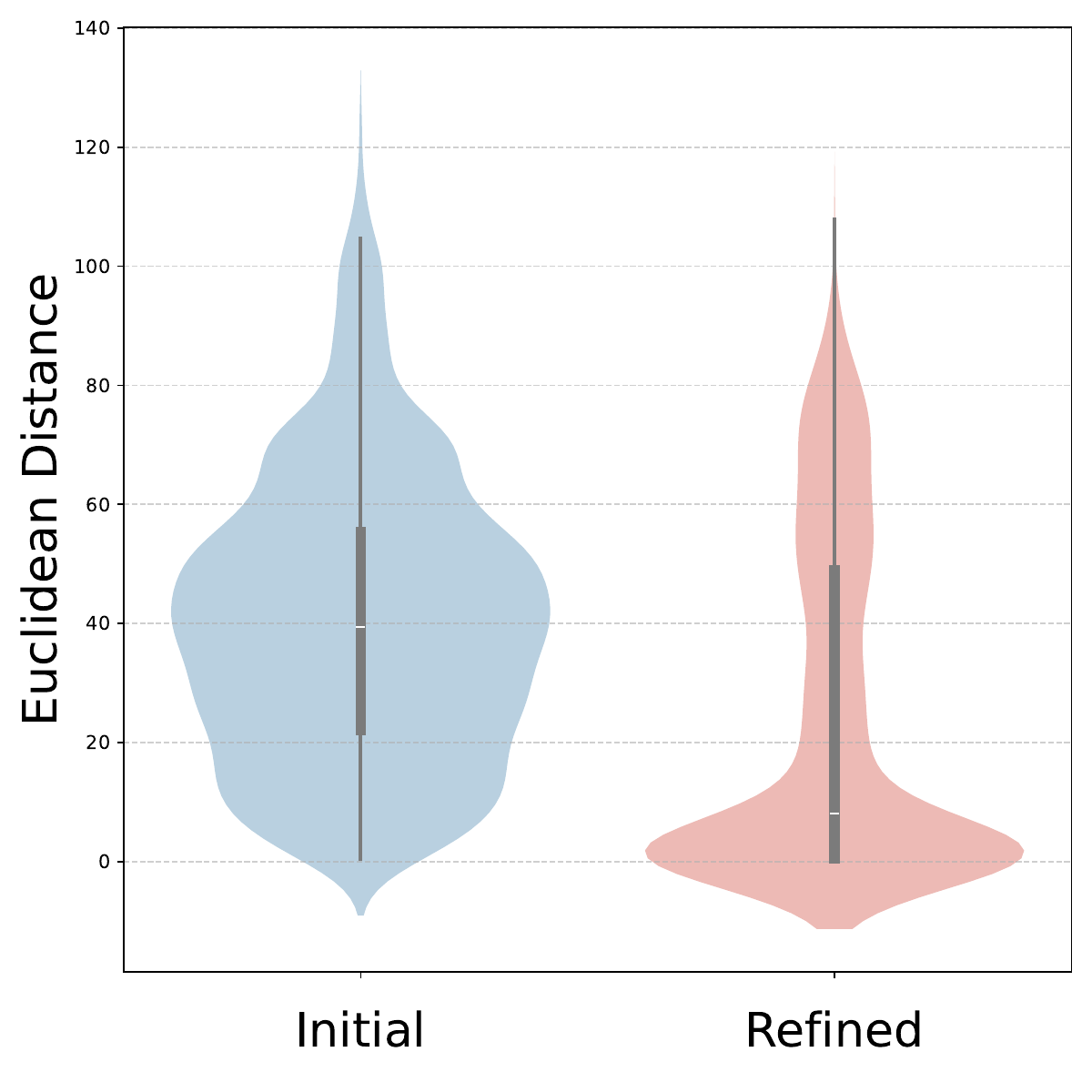}}
\end{minipage}
}
\vspace{-2mm}
\caption{\textbf{Visualization of report embeddings before and after mutual refinement.} (a) t-SNE of embedding pairs. (b) Distribution of distance between GT and generated report embeddings.}
\label{fig:point_vis}
\end{figure}
\noindent\textbf{Visualization of Stage~\uppercase\expandafter{\romannumeral2} Report Embeddings. } 
To demotrate the refinement effect of the mutual-refinement mechanism in Stage~\uppercase\expandafter{\romannumeral2}, we visualize the embeddings of the GT reports and the generated reports before and after refinement. Fig.~\ref{fig:point_vis}(a) presents the t-SNE projection of these embeddings. To focus on relative semantic distances rather than absolute positions, all GT embeddings are translated to their mean point, and the corresponding predicted embeddings are shifted by the same offset. Fig.~\ref{fig:point_vis}(b) shows the distribution of embedding distance between the generated and GT reports. It can be seen that Stage~\uppercase\expandafter{\romannumeral2} results are distributed closer to the ground truth. These two graphs illustrate that through refinement, the gap between the predicted and the GT report has been narrowed, leading to stable improvements in report quality.

\begin{figure}[!t]
\centering\includegraphics[width=\linewidth]{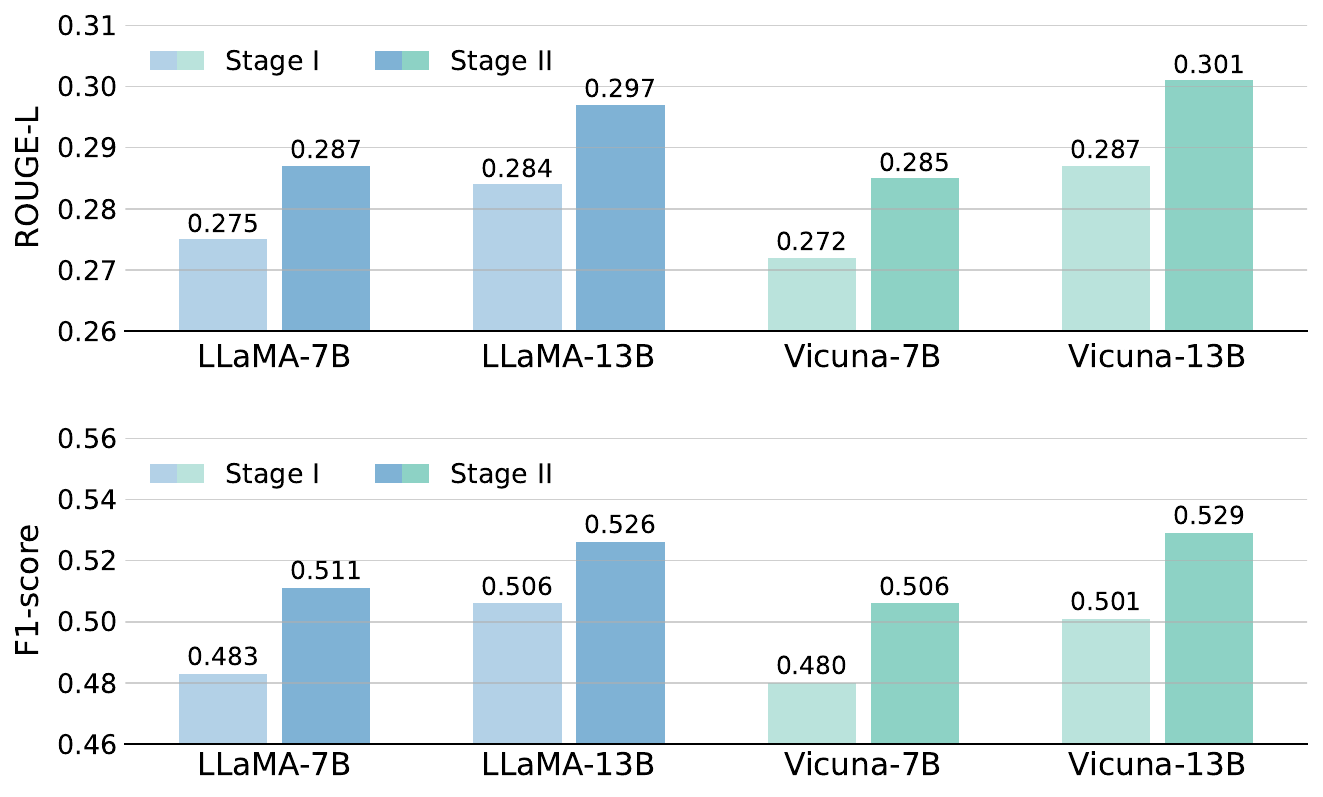}
\vspace{-2mm}
\caption{\textbf{Comparision between different LLMs.}}
\label{fig:llm_comparison}
\vspace{-3mm}
\end{figure}
\noindent\textbf{Evaluation of Cross-LLM Adaptability. }
In Fig.~\ref{fig:llm_comparison}, we evaluate the adaptability and robustness of our framework by applying four large language models in both Stage~\uppercase\expandafter{\romannumeral1} and Stage~\uppercase\expandafter{\romannumeral2}: LLaMA2-7B, LLaMA2-13B, Vicuna-7B, and Vicuna-13B. 
Results demonstrate that our approach can maintain stable performance regardless of the LLM architecture. Besides, we observe that larger LLMs achieve marginally higher scores, attributed to their superior language understanding and representation capacity.

\begin{figure}[!t]
\centering\includegraphics[width=\linewidth]{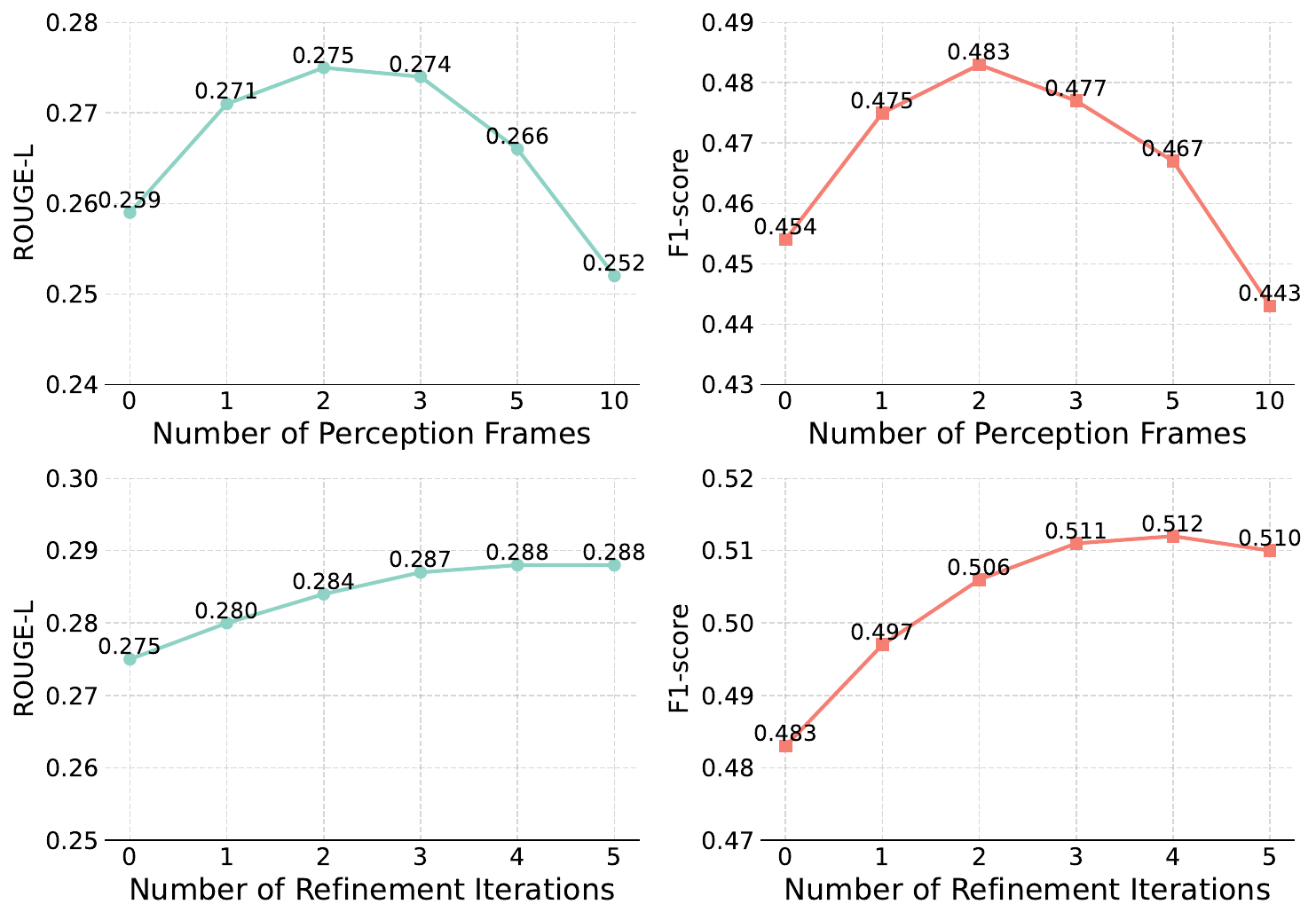}
\caption{\textbf{Hyper-parameter Analysis.} (Above) Performance with different numbers of perception frames. (Below) Results after different numbers of refinement iterations.}
\label{fig:hyperparam}
\vspace{-3mm}
\end{figure}
\subsection{Essential Hyperparameter Analysis}
\label{exp:hyperparam}
\textbf{Number of Perception Frames. }
We perform a hyperparameter analysis on the number of perception frames in Stage~\uppercase\expandafter{\romannumeral1}, evaluating performance with ROUGE-L and F1-score. Results in Fig.~\ref{fig:hyperparam} show that with no perception frames, performance drops significantly. Using 1 to 3 frames yields consistently high results, with two frames performing best. 5 or 10 frames degrade performance slightly, likely due to redundant temporal interactions. This indicates that two perception frames provide an optimal balance between temporal reasoning and model stability.

\noindent\textbf{Number of Refinement Iterations. }
We analyze the impact of refinement iterations in Stage~\uppercase\expandafter{\romannumeral2} as shown in Fig.~\ref{fig:hyperparam}. Performance improves steadily from one to three iterations, while additional iterations yield little gain, indicating that three iterations balance effectiveness and efficiency.
\section{Conclusion}
In this paper, we presented TIM, a novel framework for LRRG task. TIM decouples dynamic progression modeling from static pathological recognition and generates dual-timepoint reports to accurately capture temporal evolution. With mutual refinement, TIM effectively captures disease evolution and significantly mitigates diagnostic bias. Extensive experiments demonstrate that TIM achieves state-of-the-art results across NLG and CE metrics.
\section*{Acknowledgments.}
This project is supported by the National Key Research and Development Program of China (Grant No.2024YFE0217700), the National Natural Science Foundation of China (Grant No.62472184 and No.62122029), and the Fundamental Research Funds for the Central Universities.

{
    \small
    \bibliographystyle{ieeenat_fullname}
    \bibliography{main}
}

\clearpage
\appendix
\maketitlesupplementary
\begin{figure*}[bp]
\centering

\subfloat[]{
\begin{minipage}[]{0.24\linewidth}
\centering
\centerline{\includegraphics[width=\textwidth]{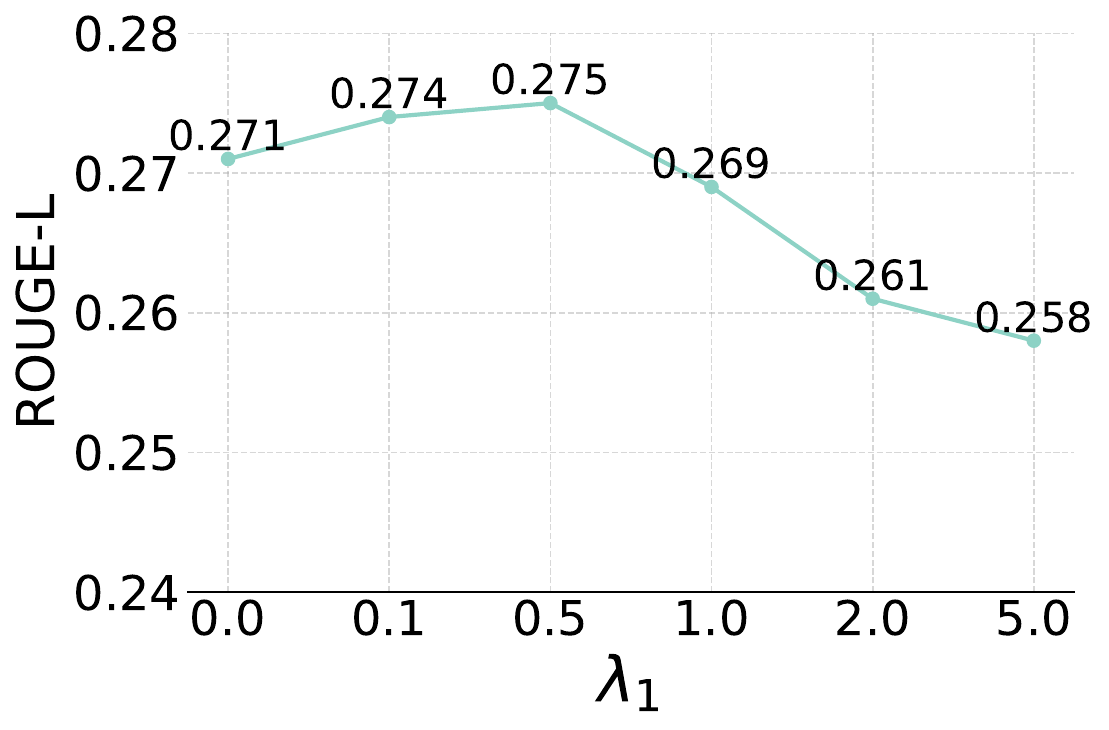}}
\end{minipage}
}
\subfloat[]{
\begin{minipage}[]{0.24\linewidth}
\centering
\centerline{\includegraphics[width=\textwidth]{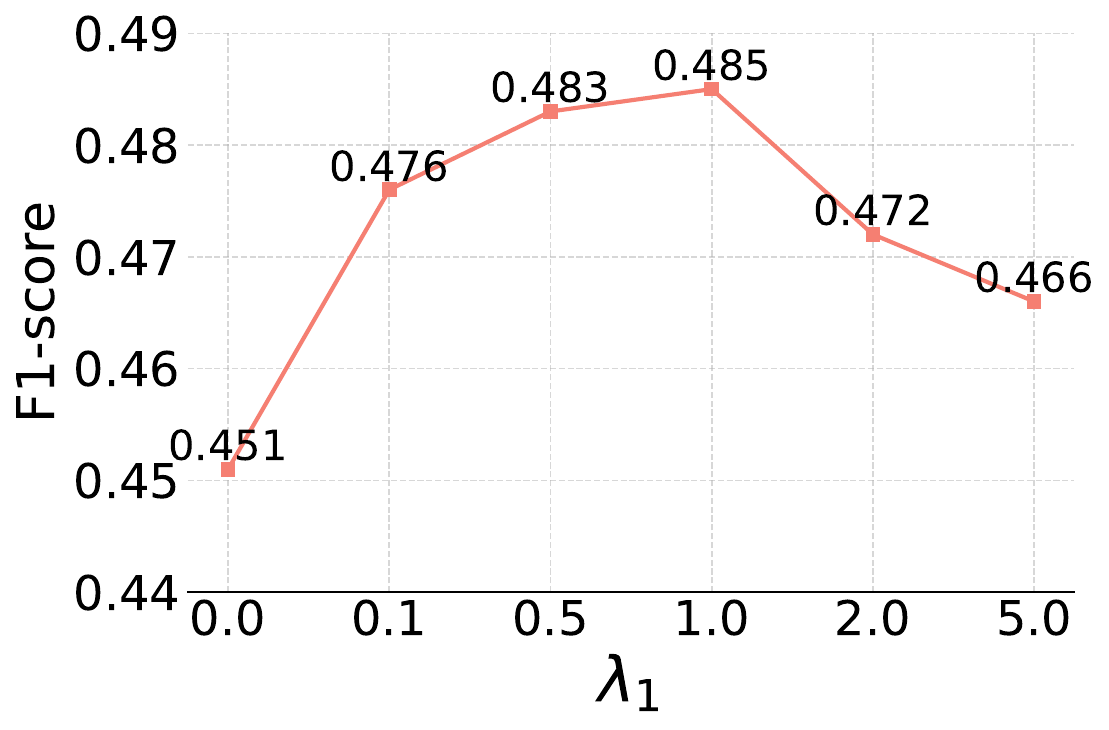}}
\end{minipage}
}
\subfloat[]{
\begin{minipage}[]{0.24\linewidth}
\centering
\centerline{\includegraphics[width=\textwidth]{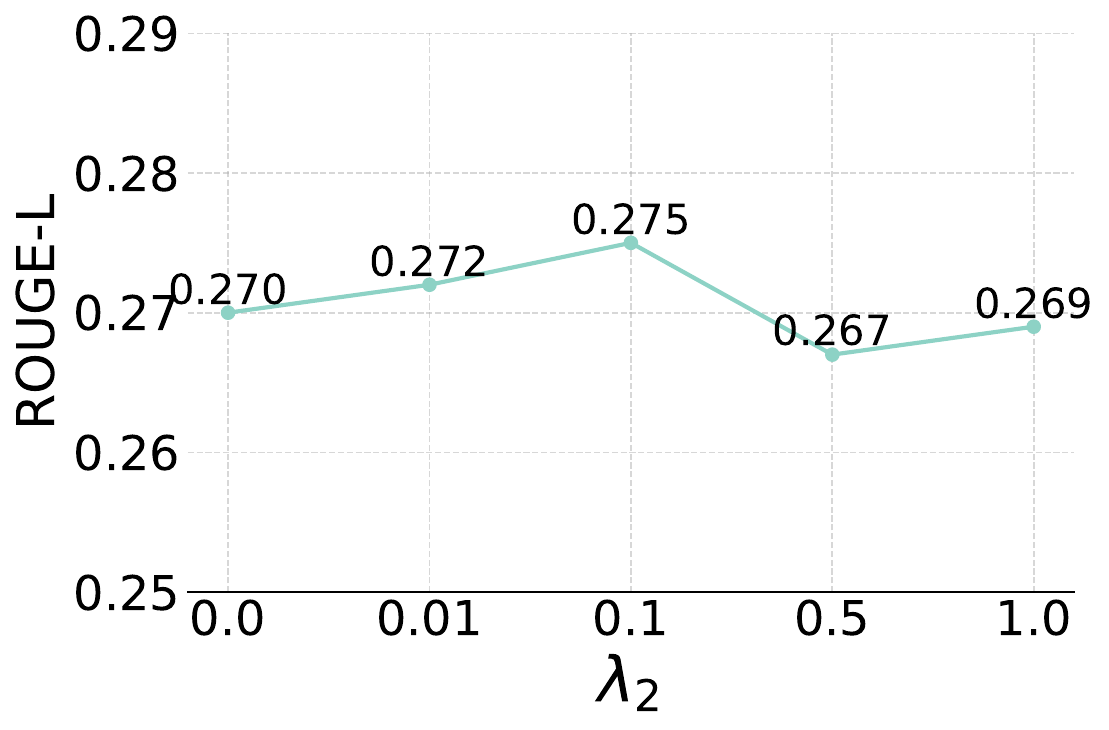}}
\end{minipage}
}
\subfloat[]{
\begin{minipage}[]{0.24\linewidth}
\centering
\centerline{\includegraphics[width=\textwidth]{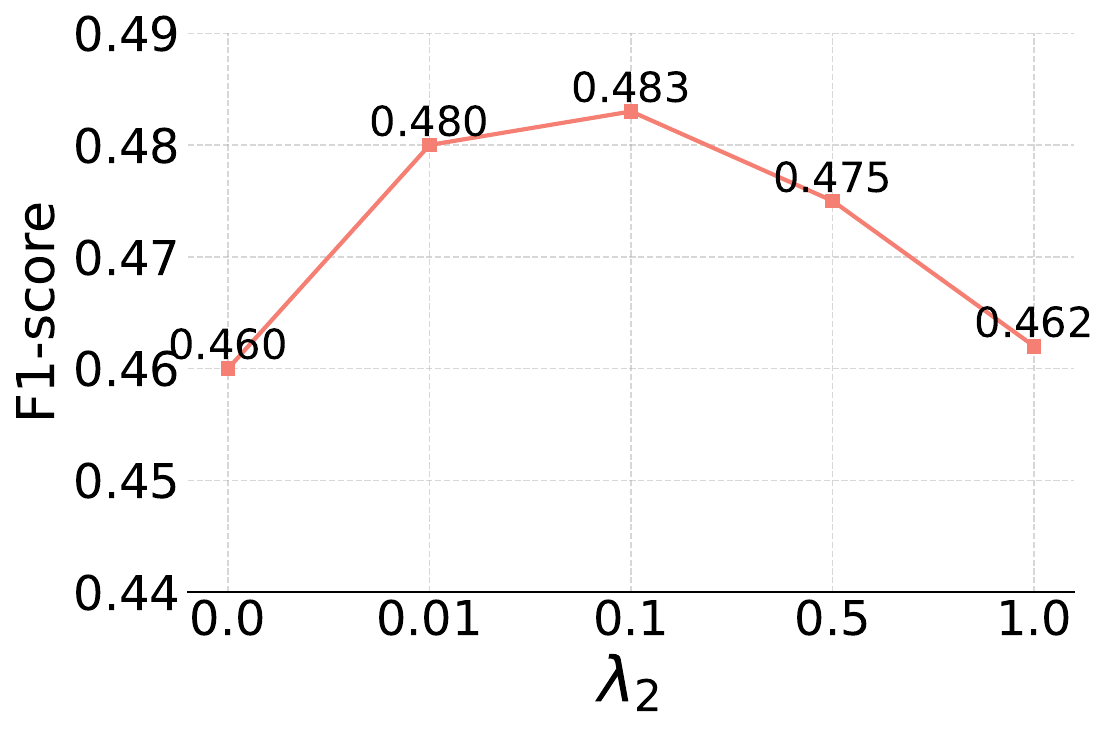}}
\end{minipage}
}
\vspace{-3mm}
\caption{\textbf{Stage I hyperparameter analysis}, including $\lambda_1$ and $\lambda_2$ in Eq.~(\ref{stage1 loss}). (a) ROUGE-L of different $\lambda_1$ values. (b) F1-score of different $\lambda_1$ values. (c) ROUGE-L of different $\lambda_2$ values. (d) F1 of different $\lambda_2$ values.}
\vspace{-3mm}
\label{fig:supp_hyperparameter_1}
\end{figure*}
\begin{figure*}[bp]
\centering

\subfloat[]{
\begin{minipage}[]{0.24\linewidth}
\centering
\centerline{\includegraphics[width=\textwidth]{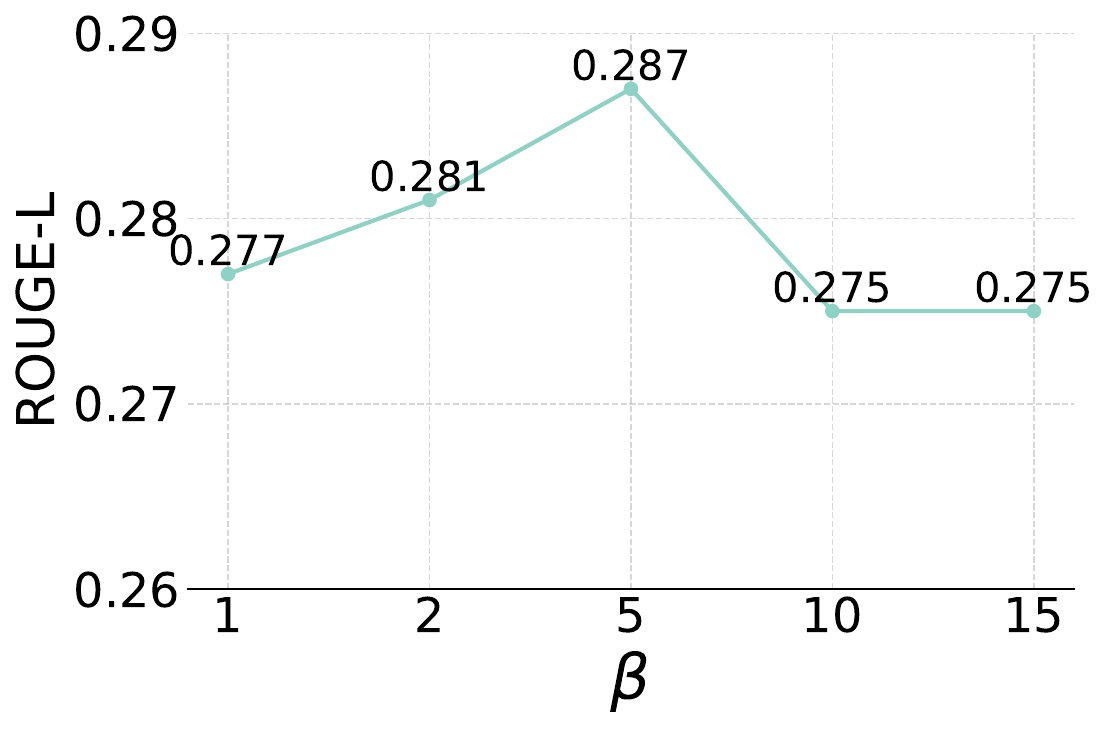}}
\end{minipage}
}
\subfloat[]{
\begin{minipage}[]{0.24\linewidth}
\centering
\centerline{\includegraphics[width=\textwidth]{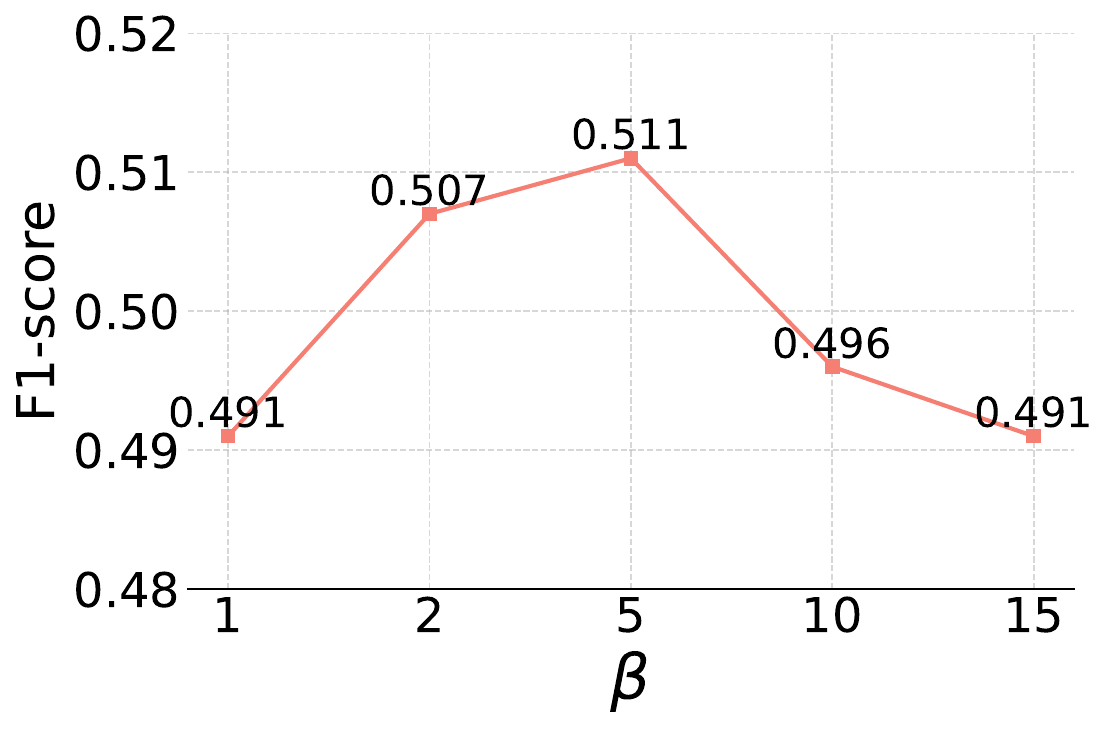}}
\end{minipage}
}
\subfloat[]{
\begin{minipage}[]{0.24\linewidth}
\centering
\centerline{\includegraphics[width=\textwidth]{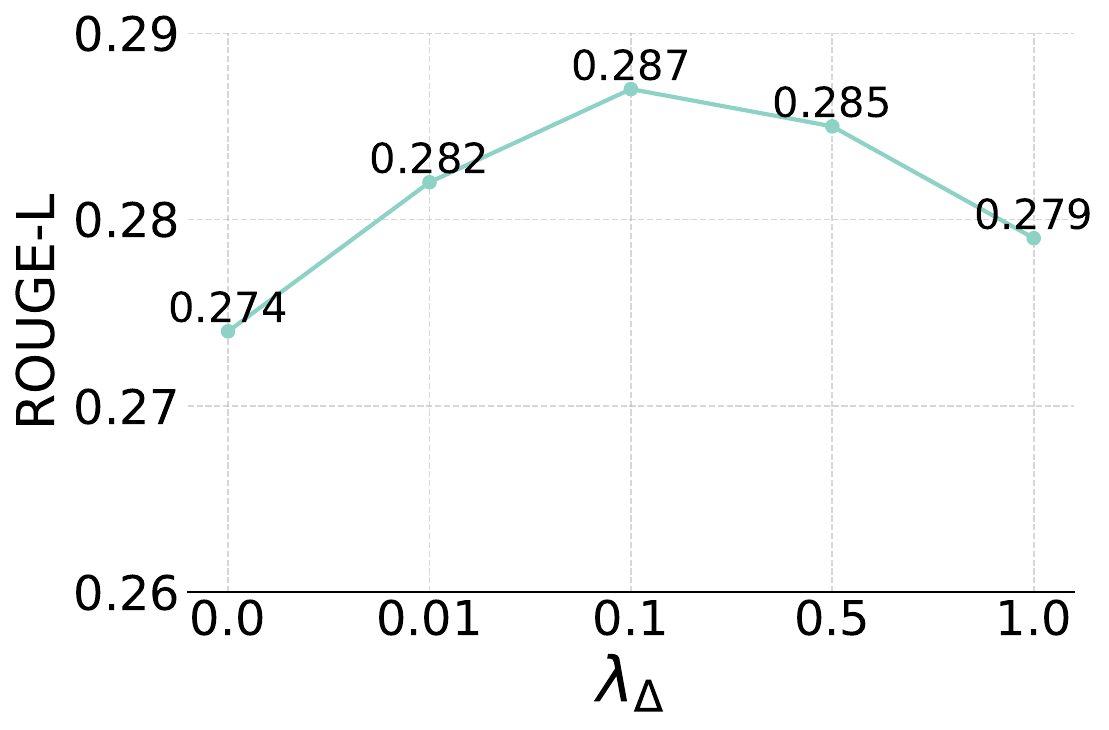}}
\end{minipage}
}
\subfloat[]{
\begin{minipage}[]{0.24\linewidth}
\centering
\centerline{\includegraphics[width=\textwidth]{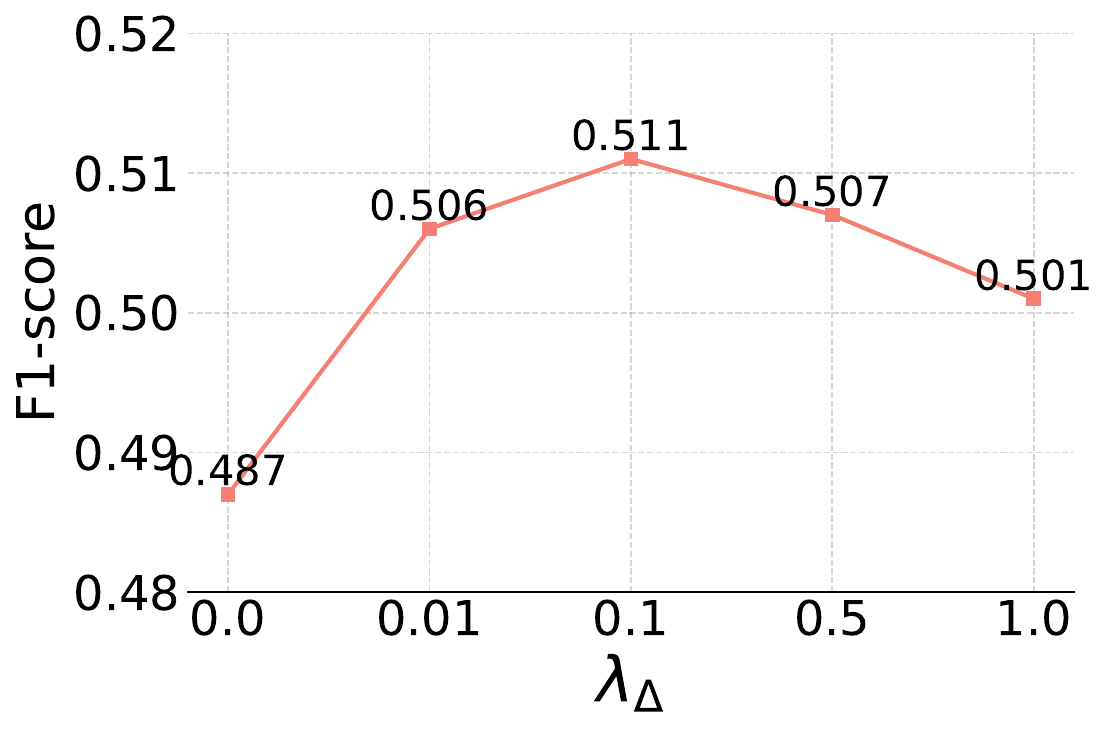}}
\end{minipage}
}
\vspace{-3mm}
\caption{\textbf{Stage II hyperparameter analysis}, including $\beta$ in Eq.~(\ref{refinement loss}) and $\lambda_\Delta$ in Eq.~(\ref{stage2 loss}). (a) ROUGE-L of different $\beta$ values. (b) F1-score of different $\beta$ values. (c) ROUGE-L of different $\lambda_\Delta$ values. (d) F1 of different $\lambda_\Delta$ values.}
\label{fig:supp_hyperparameter_2}
\vspace{-3mm}
\end{figure*}

\section{More Hyperparameter Analysis}
To further investigate the impact of hyperparameters besides the number of perception frames and the number of refinement iterations, we conducted a comparison study on four variables: the static pathological alignment weight $\lambda_1$ (cf. Eq.~(\ref{stage1 loss})), the dynamic progression alignment weight $\lambda_2$ (cf. Eq.~(\ref{stage1 loss})), the refinement scaling factor $\beta$ (cf. Eq.~(\ref{refinement loss})), and the refinement loss weight $\lambda_\Delta$ (cf. Eq.~(\ref{stage2 loss})). The experimental setup follows the dataset split in Sec.~\ref{dataset} and the implementation details in Sec.~\ref{experiment settings}. Model performance is evaluated using ROUGE-L and F1-score to measure textual fidelity and clinical accuracy.

As illustrated in Fig.~\ref{fig:supp_hyperparameter_1}, we vary $\lambda_1 \in \{0,0.1,0.5,1.0,2.0,5.0\}$ and $\lambda_2 \in \{0, 0.01, 0.1, 0.5, 1.0\}$ to study their respective roles in static pathology modeling and dynamic progression alignment. $\lambda_1=0.5$ achieves the best balance between language quality and pathology description accuracy, while $\lambda_2=0.1$ provides optimal temporal evolution modeling. Values of $\lambda_1\in[0.1,1.0]$ and $\lambda_2\in[0.01,0.5]$ yield relatively strong results, indicating the robustness of our method to moderate variations. Performance deteriorates for extreme settings, where overly small weights weaken the corresponding module contributions, while overly large weights introduce excessive alignment constraints that hinder generation quality.

For the Stage~II refinement hyperparameters , we investigate $\beta\in\{1, 2, 5, 10, 15\}$ and $\lambda_\Delta\in\{0.01, 0.1, 0.5, 1.0\}$ in Fig.~\ref{fig:supp_hyperparameter_2}. The scaling factor $\beta$ regulates the sensitivity of the sigmoid function to similarity differences, effectively controlling the sharpness of the refinement loss. The model achieves its best performance at $\beta=5$, while values within the range $\beta\in[2,10]$ also perform well. Extremely large values destabilize optimization, while smaller values diminish the corrective signal. The refinement loss weight $\lambda_\Delta$ adjusts the trade-off between the refinement objective and the autoregressive generation loss. We observe that $\lambda_\Delta=0.1$ provides the optimal balance, with values in $\lambda_\Delta\in[0.01, 0.5]$ maintaining competitive performance.

Overall, the stable performance across broad hyperparameter ranges suggests that our model is not overly sensitive to parameter tuning, making the configuration process straightforward in practice.

\section{Evaluation of Spatiotemporal Consistency}
Beyond report-level metrics, we quantitatively evaluate spatiotemporal consistency by extracting medical condition labels from predicted and ground-truth prior/current reports with CheXbert and deriving progression states (stable, worsening, improving) based on label transitions. As shown in Table.~\ref{table: temporal_comparison}, we observe that incorporating the dynamic progression modeling branch substantially improves progression F1 scores from 0.78 to 0.81, with further gains when dynamic-aware textual supervision and stage~II mutual refinement are applied, whose F1-score reaches 0.84, demonstrating more accurate temporal reasoning.
\begin{table}[h]
\caption{\textbf{Evaluation of Spatiotemporal Consistency,} conducting an ablation study on the components related to dynamic modeling.}
\centering
\renewcommand{\arraystretch}{0.9} 
\resizebox{\linewidth}{!}{%
\begin{tabular}{ccccc}
\toprule
\multirow{2}{*}{\textbf{Model}} & \multicolumn{4}{c}{\textbf{F1-score}}  \\  \cmidrule{2-5} 
                                & \multicolumn{1}{l}{\textbf{Stable}} & \multicolumn{1}{l} {\textbf{Worsening}} & \multicolumn{1}{l}{\textbf{Improving}} & \multicolumn{1}{l}{\textbf{Avg.}}\\ \midrule 
TIM w/o Progression Branch & 0.85 & 0.30 & 0.52 & 0.78\\
TIM w/o Progression Loss   & 0.89 & 0.31 & 0.58 & 0.81\\
TIM-stage I                & 0.89 & 0.33 & 0.61 & 0.82 \\
TIM-stage II               & \textbf{0.91} & \textbf{0.35} & \textbf{0.65} & \textbf{0.84}\\
\bottomrule
\end{tabular}%
}
\label{table: temporal_comparison}
\end{table}

\section{Example of Prompts for LLM Input}
In both Stage I and Stage II, the LLM receives a textual prompt that guides the report generation process. Each prompt specifies the model's role as a radiologist, outlines the task and the expected input-output behavior, and indicates the temporal context of the report (i.e., prior or current examination). Below, we provide examples of the prompts used in two stages. The symbol ``[$\cdot$]'' denotes the corresponding embedding inputs:

\begin{tcolorbox}[title = {Stage I Input Prompts}]
\emph{\textless s\textgreater{} User: Progression: [progression]. Prior report: [Context]. Image: [Image]. You are a radiologist. Please generate a radiology report for the provided current chest X-ray image, using the visual findings from the image, the text of the previous report, and the progression between current and prior images. Assistant:}
\end{tcolorbox}

\begin{tcolorbox}[title = {Stage II Input Prompts}]
\emph{\textless s\textgreater{} User: You are a radiologist. You have already written reports for the current and previous chest X-ray images of a patient, but there might be some issues in those reports. Current image: [Image]. Your report: [Report]. The following pathology was not consistent with the actual report in the generation of your previous report: [Diff]. Please pay special attention to these errors and regenerate an accurate report. Assistant:}
\end{tcolorbox}

\section{Examples of Clinical Descriptions of Medical Conditions}

As demonstrated in Section~\ref{sec: pathological recognition}, to couple pathological features with clinical semantics, we expand each medical condition into a clinical description using external medical knowledge sources such as UMLS and PubMed. Each description summarizes the characteristic radiographic manifestation of the disease on chest radiographs and follows a standardized template in the form of ``\{\emph{condition}\} on chest x-ray refers to ...'' to ensure consistent linguistic structure across disease categories. This formulation preserves key clinical terminology (e.g., opacity, cardiomegaly, air bronchogram) while maintaining concise semantics suitable for vision–language alignment. Table~\ref{tab:medical conditions} presents representative examples of these disease-aware textual descriptions.

\section{Case Studies}

\begin{figure*}[bp]
\centering\includegraphics[width=\linewidth]{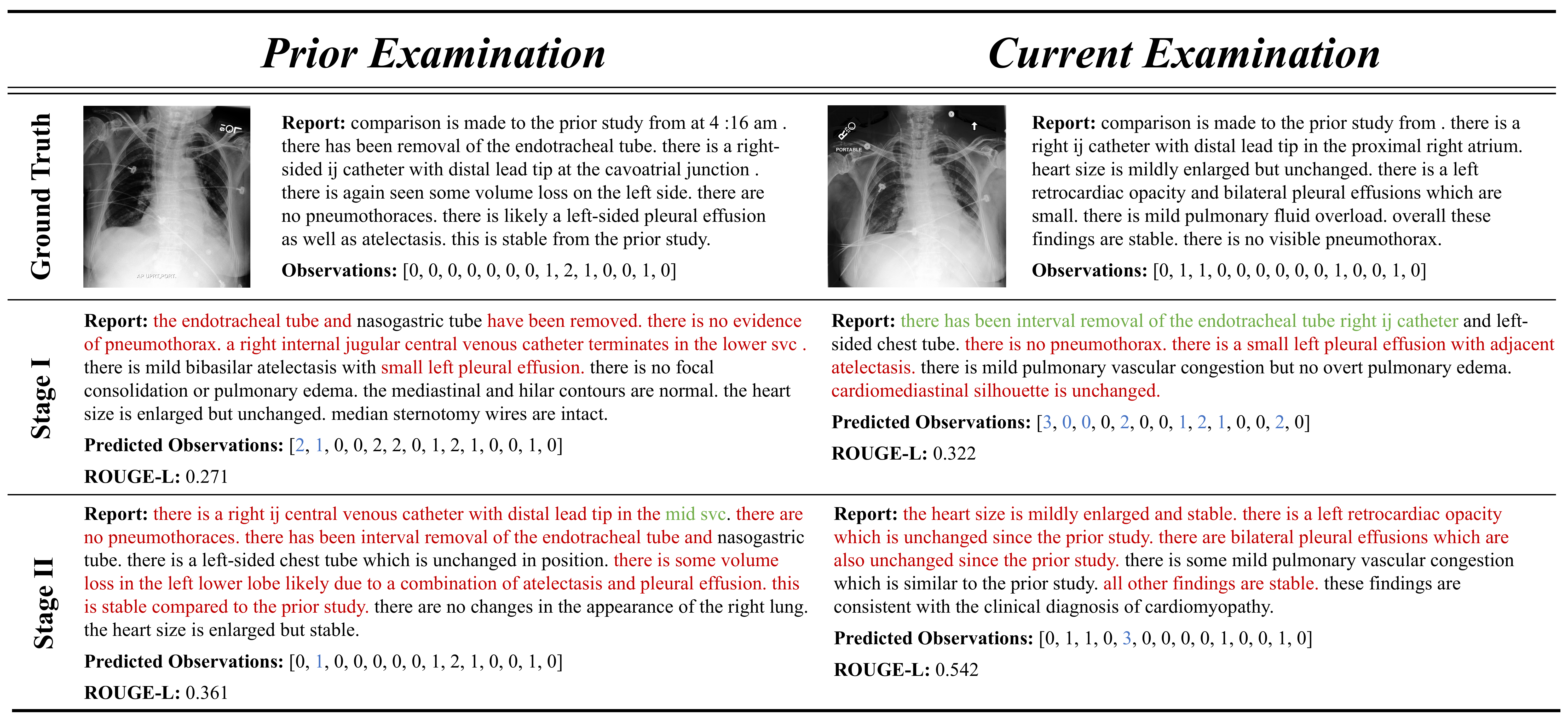}
\caption{\textbf{Specific Case Study 1.} 
\textcolor{deep_red}{Red} and \textcolor{shallow_green}{green} fonts indicate statements that are \textcolor{deep_red}{consistent} with and \textcolor{shallow_green}{conflicting} with the ground truth, respectively. \textcolor{deep_blue}{Blue} numbers in the predicted observations mark classification errors.}
\label{fig:supp_case_1}
\end{figure*}

To further illustrate the effectiveness of our proposed TIM, we present four case studies. Each case compares model outputs from both Stage I and II with the ground truth reports, shows the predicted observations, and reports the ROUGE-L score for each output. The 14 observation categories are \emph{\{enlarged cardiomediastinum, cardiomegaly, lung opacity, lung lesion, edema, consolidation, pneumonia, atelectasis, pneumothorax, pleural effusion, pleural other, fracture, support devices, no finding\}}. The label codes 0, 1, 2, and 3 denote \emph{\{not mentioned, positive, negative, uncertain\}}.

\begin{figure*}[!bp]
\centering\includegraphics[width=\linewidth]{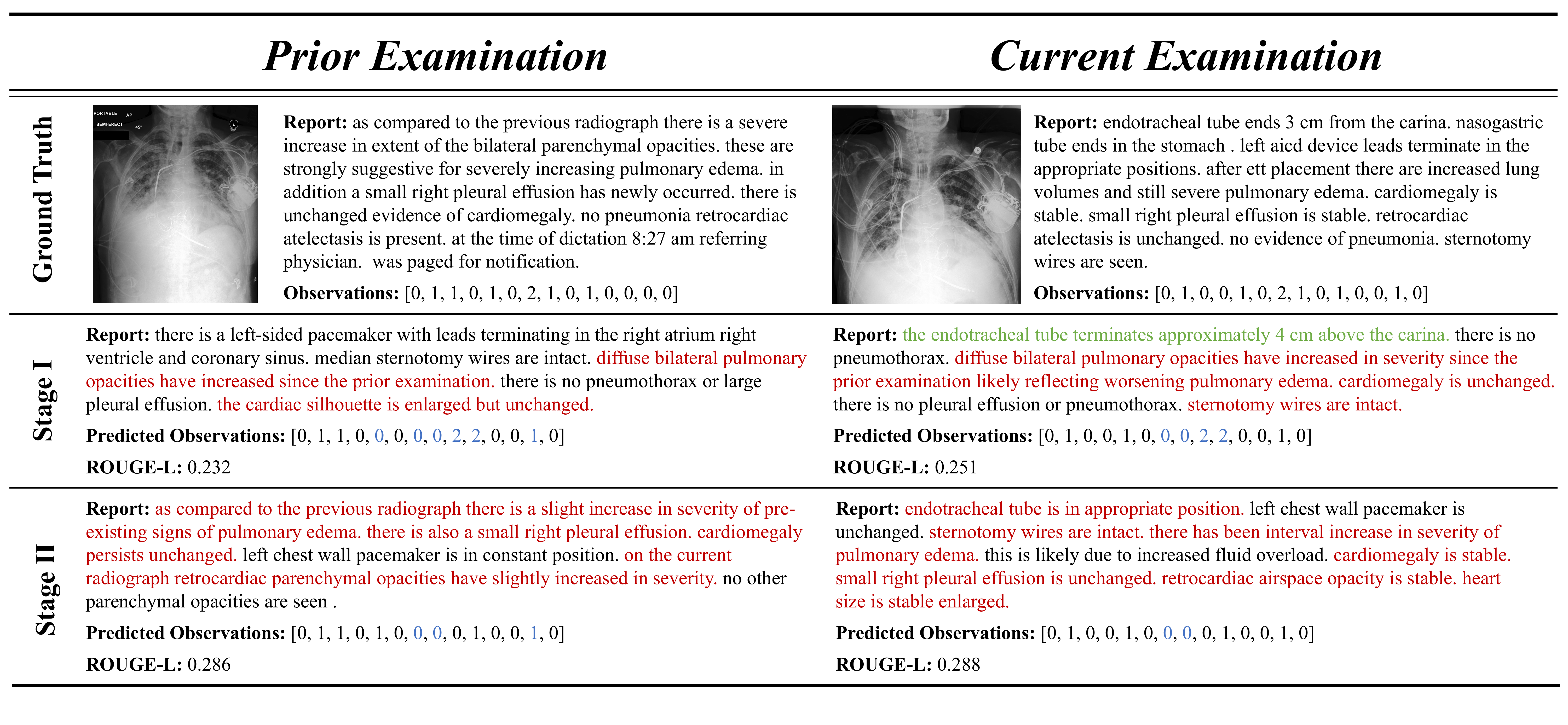}
\caption{\textbf{Specific Case Study 2.} 
\textcolor{deep_red}{Red} and \textcolor{shallow_green}{green} fonts indicate statements that are \textcolor{deep_red}{consistent} with and \textcolor{shallow_green}{conflicting} with the ground truth, respectively. \textcolor{deep_blue}{Blue} numbers in the predicted observations mark classification errors.}
\label{fig:supp_case_2}
\end{figure*}

\begin{figure*}[!bp]
\centering\includegraphics[width=\linewidth]{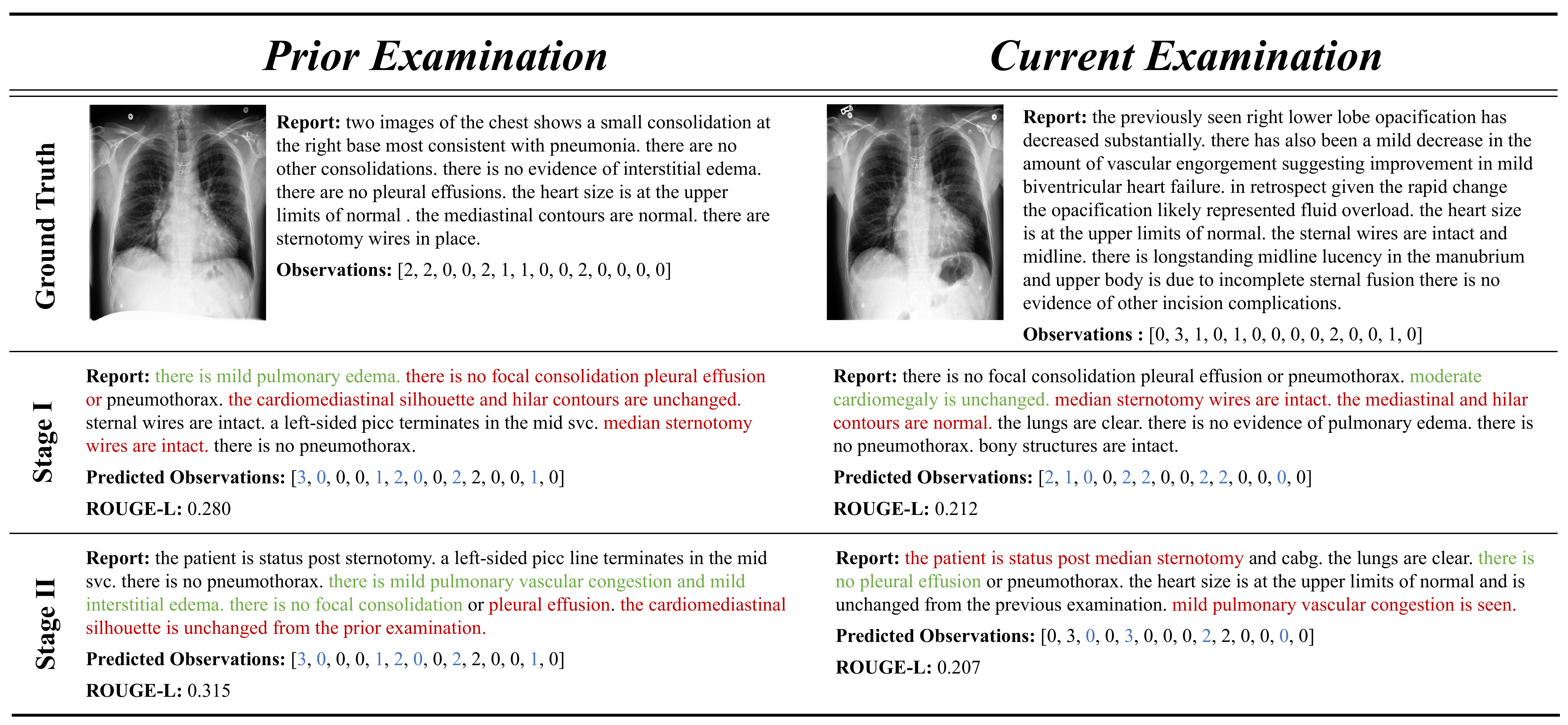}
\caption{\textbf{Specific Case Study 3.} 
\textcolor{deep_red}{Red} and \textcolor{shallow_green}{green} fonts indicate statements that are \textcolor{deep_red}{consistent} with and \textcolor{shallow_green}{conflicting} with the ground truth, respectively. \textcolor{deep_blue}{Blue} numbers in the predicted observations mark classification errors.}
\label{fig:supp_case_4}
\end{figure*}

As shown in Figs.~\ref{fig:supp_case_1}--\ref{fig:supp_case_4}, these qualitative evaluations demonstrate how our method captures temporal disease progression and corrects cross-time inconsistencies. In the predicted reports, red and green fonts indicate statements that are consistent with and conflicting with the ground truth, respectively. Blue numbers in the predicted observations mark classification errors. 

Case 1 is shown in Fig.~\ref{fig:supp_case_1}. For the prior exam, both the ground truth and Stage I agree that the endotracheal tubes are removed, and no pneumothorax is present. Stage II preserves these correct findings, but also introduces a left chest tube that is not described in the prior ground truth. For the current exam, Stage I incorrectly omits the right IJ catheter described in the ground truth and reports removal of the chest tube, which contradicts the ground truth. Stage II corrects most of these errors for the current study: it restores the presence and stability of the left retrocardiac opacity, explicitly reports bilateral small pleural effusions and mild pulmonary vascular congestion, and states that heart size is mildly enlarged but stable, all consistent with the ground truth. 

Case 2 is shown in Fig.~\ref{fig:supp_case_2}. For the prior examination, Stage I correctly reports increased bilateral pulmonary opacities and cardiomegaly but omitted the newly developed small right pleural effusion described in the ground truth. Stage II recovers the pleural effusion and preserves the finding of unchanged cardiomegaly. For the current examination, Stage I captures worsening bilateral opacities and stable cardiomegaly, but again misses the small right pleural effusion and provides a slightly different endotracheal tube position. After refinement, Stage II restores the small right pleural effusion and retrocardiac air-space opacity reported in the ground truth. 


From these cases, we observe that Stage II refinement improves both report completeness and disease-label accuracy compared with Stage I outputs, indicating that the refinement process reduces repeated errors across timepoints and yields reports that align more closely with the ground truth progression.

We also present a representative failure case in which the model does not recover the subtle temporal evolution described in the ground truth reports. As shown in Fig.~\ref{fig:supp_case_4}, the prior ground truth report documents a small right-basilar consolidation, and the current GT report describes a rapid interval decrease of this opacity. However, the visual appearance of this lesion is extremely subtle and often confounded by postoperative changes, vascular markings, and acquisition variability across the two time points. As a result, both Stage I and Stage II tend to produce more conservative normal findings, which is a common behavior observed when the radiographic evidence for a minor abnormality is ambiguous. This example illustrates a challenging scenario driven by imaging subtlety. Future work may incorporate uncertainty modeling to better handle such borderline cases.


\begin{table*}[t]
\centering
\caption{Examples of clinical descriptions used for disease-aware text construction.}
\renewcommand{\arraystretch}{1.3}
\begin{tabularx}{\textwidth}{m{3cm}<{\centering} m{3.2cm}<{\centering}  m{10.2cm}}
\toprule
\textbf{Medical Condition} & \textbf{Source Literature} & \makecell[c]{\textbf{Annotation Prompt}} \\
\midrule

\textbf{Enlarged Cardiomediastinum} & Klein et al.~\cite{Rosado-de-Christenson2015} &
Enlarged cardiomediastinum on chest x-ray refers to the widening or abnormal enlargement of the combined cardiac and mediastinal contours on chest radiography. \\

\textbf{Cardiomegaly} & Siwik et al.~\cite{arm91020010}  Cardinale et al.~\cite{Cardinale2012CXR} &
Cardiomegaly on chest x-ray refers to an abnormally enlarged cardiac silhouette, typically identified when the cardiac-thoracic ratio (CTR) exceeds 0.5, meaning the heart width occupies more than half of the thoracic cage width. \\

\textbf{Lung Opacity} & Turk et al.~\cite{Turk2023LungOpacity} &
Lung opacity on chest x-ray refers to an area in the normally dark lung that appears whiter, hazy, or denser due to increased tissue density from the replacement of air by substances such as fluid, fibrosis, or other material. \\

\textbf{Lung Lesion} & Feyisa et al.~\cite{Feyisa2023WeakLocalizationTB} &
Lung lesion on chest x-ray refers to pulmonary lesions presenting as localized nodules, masses, or cavitary abnormalities within the lung parenchyma, often associated with structural distortion of surrounding lung tissue. \\

\textbf{Edema} & Siwik et al.~\cite{arm91020010} &
Pulmonary edema on chest x-ray refers to interstitial or alveolar fluid accumulation, often producing distortion in the hilar image, thickening of interlobar or interlobular fissures, and bilateral higher density opacities in central and basal lung areas. \\

\textbf{Consolidation} & Bankier et al.~\cite{Bankier2024FleischnerGlossary} Cardinale et al.~\cite{Cardinale2012CXR} &
Consolidation on chest x-ray refers to a homogeneous increase in attenuation caused by alveolar filling with fluid, pus, blood, or cells, resulting in homogeneous opacification of the lung parenchyma on radiographs, often accompanied by air bronchograms. \\

\textbf{Pneumonia} & Bankier et al.~\cite{Bankier2024FleischnerGlossary} &
Pneumonia on chest x-ray refers to infection-related lung abnormalities that manifest as opacity, potentially accompanied by air bronchograms (air bronchogram), tree-in-bud opacities, or pleural effusions (pleural effusion). \\

\textbf{Atelectasis} & Klein et al.~\cite{Rosado-de-Christenson2015} &
Atelectasis on chest x-ray refers to partial or complete collapse of lung tissue leading to volume loss, increased density on radiographs, and displacement of adjacent structures such as fissures or mediastinum. \\

\textbf{Pneumothorax} & Bankier et al.~\cite{Bankier2024FleischnerGlossary} &
Pneumothorax on chest x-ray refers to the presence of pleural air typically identified on upright chest radiographs by a curvilinear visceral pleural line indicating the demarcation between pleural air and aerated lung. \\

\textbf{Pleural Effusion} & Bankier et al.~\cite{Bankier2024FleischnerGlossary} &
Pleural effusion on chest x-ray refers to accumulation of fluid in the pleural space where small pleural effusions manifest as blunting of the costophrenic angles, often meniscus-shaped, and larger effusions as homogeneous opacities in dependent regions. \\

\textbf{Pleural Other} & Bankier et al.~\cite{Bankier2024FleischnerGlossary} Eisenhuber et al.~\cite{Eisenhuber2012BedsideCXR} &
Pleural other on chest x-ray refers to other pleural abnormalities presenting as pleural thickening, plaques, or irregular pleural contours, sometimes associated with calcifications or localized pleural masses. \\

\textbf{Fracture} & Dilday et al.~\cite{Dilday2024RibFractures} &
Fracture on chest x-ray refers to rib fractures appearing as cortical discontinuities or step deformities along the rib contour and may be associated with complications such as pneumothorax or hemothorax. \\

\textbf{Support Devices} & Hunter et al.~\cite{Hunter2004ChestDevices} Baratella et al.~\cite{Baratella2021ICUDevices} &
Support devices on chest x-ray refers to medical support devices such as endotracheal tubes, central venous catheters, pacemaker leads, and thoracostomy tubes appearing as radiopaque linear or tubular structures used to confirm correct positioning and detect potential complications. \\

\bottomrule
\end{tabularx}
\label{tab:medical conditions}
\end{table*}


\end{document}